\documentclass[twocolumn]{Styles/autart}  % Comment this line out if you need a4paper

\makeatletter
\let\NAT@parse\undefined
\makeatother
\usepackage{mathptmx}       % selects Times Roman as basic font
\usepackage{makeidx}         % allows index generation
\usepackage{graphicx,graphbox}        % standard LaTeX graphics tool
\usepackage{multicol}        % used for the two-column index
\usepackage{amsmath,amssymb,wrapfig}
\usepackage{euscript}
\usepackage{mathrsfs}
\usepackage{todonotes}
\usepackage{array}
\usepackage{gatechamberlab}
\usepackage{epstopdf}
\usepackage{booktabs}
\usepackage{multirow}
\usepackage{subfig}
\usepackage{tikz}
\usepackage{color}
\usepackage{arydshln}\usepackage[normalem]{ulem}
\usepackage[colorlinks]{hyperref}
\hypersetup{
    colorlinks,
    linkcolor=blue,
    citecolor=blue,
    filecolor=magenta,
    urlcolor=cyan,
}

\newcommand{\addblue}[1]{\textcolor{black}{#1}}

\begin{document}

% \onecolumn
% \input{Reviews/Reviewer}

\twocolumn
% \newrefcontext 

\begin{frontmatter}
	
	\title{Local Stability of PD Controlled Bipedal \addblue{Walking Robots}
		\thanksref{footnoteinfo}} % Title, preferably not more 
	% than 10 words.
	
	\thanks[footnoteinfo]{This work is supported by the DST INSPIRE Faculty Fellowship IFA17-ENG212 and the Robert Bosch Center for Cyber Physical Systems. Shishir Kolathaya is an INSPIRE Faculty Fellow at the Center for Cyber Physical Systems, Indian Institute of Science, Bengaluru. Corresponding author Shishir Kolathaya. Tel. +91 80 2360 0644. 
		}
        
	\author[Bengaluru]{Shishir Kolathaya}\ead{shishirk@iisc.ac.in}    % Add the 
	
	\address[Bengaluru]{Center for Cyber Physical Systems, Indian Institute of Science, Bengaluru, Karnataka, India}  % Please supply

	                             % keyword list or with the 
	% help of the Automatica 
	% keyword wizard

\begin{abstract}

We establish stability results for PD tracking control laws in bipedal walking robots.
Stability of PD control laws for continuous robotic systems is an established result, and we extend this for hybrid robotic systems, 
an alternating sequence of continuous and discrete events.
Bipedal robots have the leg-swing as the continuous event, and the foot-strike as the discrete event. In addition, bipeds largely have underactuations due to the interactions between feet and ground.
For each continuous event, we establish that the convergence rate of the tracking error can be regulated via appropriate tuning of the PD gains; and for each discrete event, we establish that this convergence rate sufficiently overcomes the nonlinear impacts 
by assumptions on the hybrid zero dynamics.
The main contributions are 1) Extension of the stability results of PD control laws for underactuated robotic systems, and 2) Exponential ultimate boundedness of hybrid periodic orbits under the assumption of exponential stability of their projections to the hybrid zero dynamics. Towards the end, we will validate these results in a $2$-link bipedal walker in simulation.
\end{abstract}

\end{frontmatter}

\section{Introduction}

Despite great advances in the theory of nonlinear controls, when it comes to practical implementation, PD control laws undisputably continue to be the most popular choice. \cite[Table 1A]{7823045} shows a detailed account on the list of controllers used and their corresponding acceptance ratings. The popularity of this type of control laws arises from its ease of implementation and robustness due to its model independent nature. %, and versatility due to its robustness to uncertainties. 
This popularity has equally pervaded %PD based control laws are ubiquitous in 
robotic systems. In fact, there are formal guarantees of stability for a broad class of robots that includes manipulators \cite{koditschek-the_robotics_review-1989,wen1988new}. See Table \ref{tab:formalresults}, which shows a list of stability results. A more detailed list of these types of control laws and the corresponding stability results are given in \cite[Table 1.1]{choi2004pid}.

\begin{figure}[!ht]\centering
	\includegraphics[height=0.53\columnwidth]{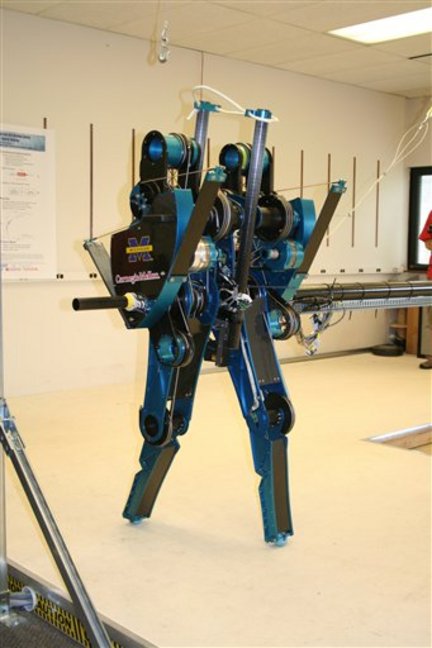}\hspace{-1mm}
	\includegraphics[height=0.53\columnwidth]{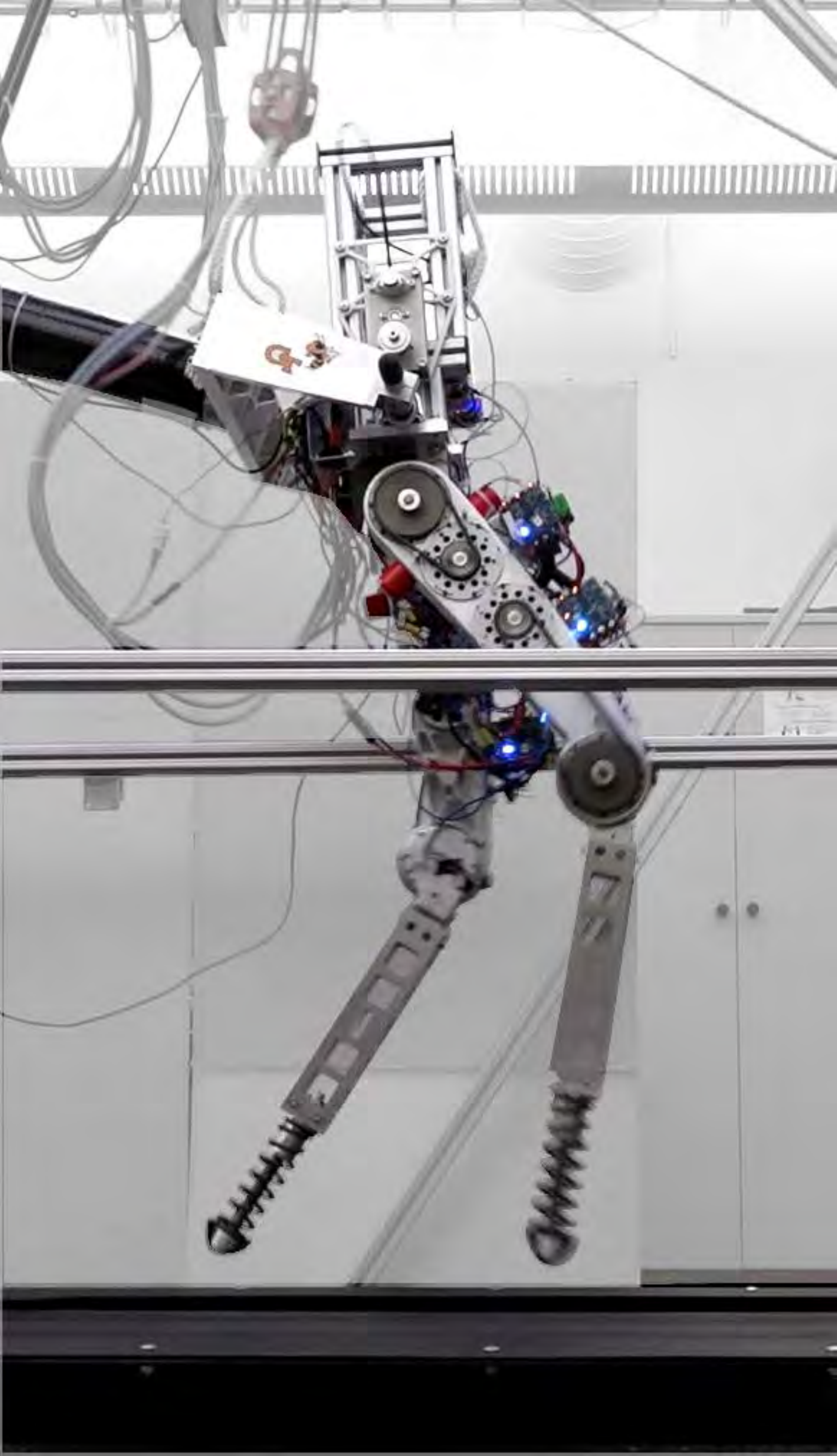}\hspace{-1mm}
	\includegraphics[height=0.53\columnwidth]{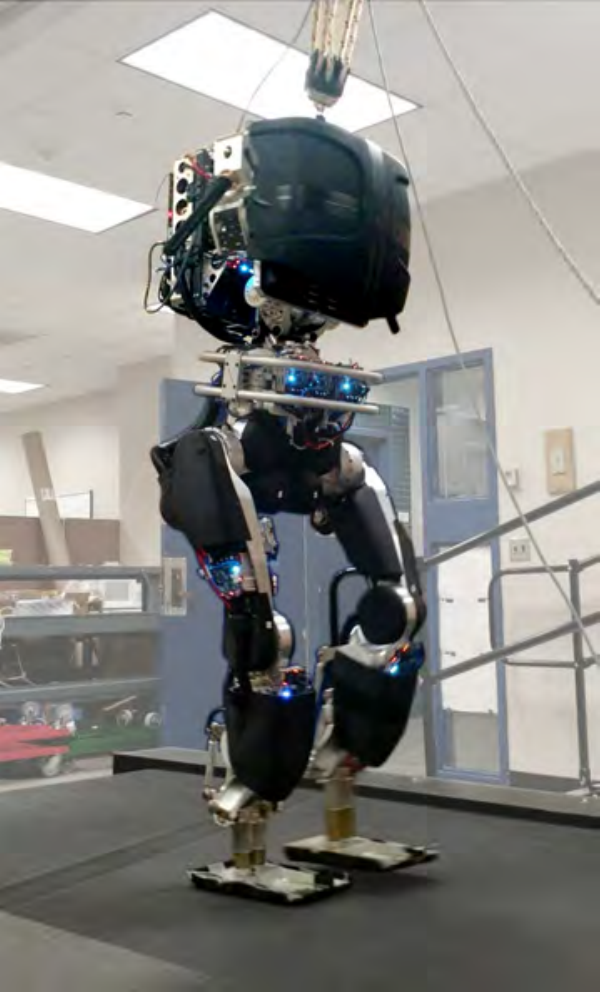}	%\includegraphics[width=\textwidth]{Figures/robotlistPD.pdf}
	\caption{
		Figure showing the MABEL walking robot \cite{umich_mabel} (left), DURUS-2D running robot \cite{hscc17running} (middle), and the DURUS walking robot \cite{kolathaya2016time,reheralgorithmic} (right). All of these bipeds used PD control laws for tracking reference trajectories at the low level. 
	}
	\label{fig:robotlistPD}
\end{figure}

	\begin{table}[!ht]
\begin{center}
		\begin{tabular}{|l | l |}
			\hline
% 			PD regulation  &  UB \cite{12253} \\
% 			\hline 
			PD regulation  & GAS \cite{koditschek1984natural,takegaki1981new} \\
% 			+ Gravity compensation 
			& LES \cite{ARIMOTO1985221,koditschek1987adaptive}, GES  \cite{koditschek1988global} \\
			\hline 
% 			PD tracking		&  UUB \cite{12253} \\
% 			\hline
			PD tracking  & LES \cite{12253,koditschek1987highgain,wen1988new,whitcombpd1993} \\
% 			+ Gravity compensation 
% 			&  \\
			\hline
		\end{tabular}
		\caption{Table showing some of the formal stability results for PD and PD based control laws for robotic systems. The abbreviations are given as follows: %UB: ultimately bounded, LEUB: locally exponentially ultimately bounded, 
		GAS: globally asymptotically stable, LES: locally exponentially stable, GES: globally exponentially stable.
% 		ISS: input-to-state stable, UUB: uniformly ultimately bounded.
		}
		\label{tab:formalresults}
\end{center}
	\end{table}

Despite the increase in complexity of the models, PD control laws have dominated even the domain of bipedal robots. Bipedal locomotion is hybrid in nature, with alternating phases of continuous (swinging forward of the nonstance foot) and discrete events (instantaneous impact of the nonstance foot with ground). In addition, unlike the industrial robotic arms, which have a fixed base, bipedal robots are largely underactuated. %due to the fact that the stance foot is not pinned to ground. 
%This is due to the fact that the stance foot is not pinned to ground.
Fig. \ref{fig:robotlistPD} shows some examples of bipedal robots that used PD and PD based tracking control laws. 
It is worth noting that the reference trajectories that were tracked were obtained offline.
In fact, the field of locomotion has largely focused on experimental realization by dividing the problem into two parts. First, obtain reference trajectories/gaits in a simulation model by using offline optimization tools \cite{Hereid_etal_2016,posa2014direct}. Second, play these trajectories in the robot by a low level tracking control law \cite{Hubicki2016,reheralgorithmic,Yadukumar2013}. Due to uncertainties in the system, model based controllers were generally avoided, thereby giving preference to the more traditional PD based control laws. These control laws are known to be ``hassle free", since they are model-independent and easy to implement. %, and the increasing torque densities of BLDC drives enabled better tracking performances with this type of control methodology. 
Therefore, the main goal of this paper is to explore the stability properties of PD control laws for walking robots, which include varying levels of complexity due to the presence of impacts and underactuations. 

We will be establishing stability of PD control laws by the construction of strict Lyapunov functions developed by Arimoto et. al. \cite{ARIMOTO1985221}, Koditschek \cite{koditschek1987adaptive}, and Bayard and Wen \cite{wen1988new} all in the same period of time 1984-1988. Local and global stability results were shown for both stationary and time varying desired configurations \cite{12253,koditschek1987highgain,whitcombpd1993}, but only for fully actuated systems. %These results are well known for fully actuated systems, but cannot be directly extended to underactuated systems. 
For underactuated systems, we can apply a tracking control law for the actuated states of the robot, but this does not necessarily guarantee stability, due to the coupling between the controllable and uncontrollable dynamics of the robot. In addition, the impacts due to foot-strike largely have ``destabilizing" effects on the tracking errors. On the other hand, if we make assumptions about the uncontrolled dynamics of the robot i.e., existence of stable periodic orbits in the hybrid zero dynamics (HZD), we can then establish local stability results. %This assumption on the HZD is very well known in the walking community \cite{grizzle20103d}, and have been successfully used to realize both walking and running. 
This will be the main approach of the paper.

%\newsec{Hybrid Zero Dynamics based control.} 
It is important to discuss the notion of hybrid zero dynamics (HZD) in the context of bipedal walking. HZD was introduced as a feedback design method to move beyond the traditional quasi-static flat-footed walking gaits \cite{grizzle20103d,WGK03}. The goal was to realize a stable walking gait (periodic orbit) via stabilization of a subset of the states of the robot, while the remaining states of the system exhibit uncontrolled dynamics (resulting in HZD). It was shown in \cite{WGK03} that if there is an exponentially stable periodic orbit in the HZD, then by employing a suitable output stabilizing control law \cite{TAC:amesCLF,MOGR05}, the periodic orbit of the full hybrid dynamics can be stabilized, resulting in stable walking.

There were two main output stabilizing controllers proposed over the last ten years using this notion of HZD---feedback linearization \cite{MOGR05} and control Lyapunov functions (CLF) \cite{TAC:amesCLF,kolathaya2016time}. Both of these control methodologies relied on using a user defined $\epsilon$, which was, in principle, increasing the controller gain. With a sufficiently large gain, the destabilizing impact events were overcome by faster convergences of the outputs. Our focus in this paper is to realize the same behavior via PD based output stabilizing control laws, wherein the desirable convergence rates are obtained via tuning of the PD gains.

Despite their widespread use in practical robotic systems, PD control laws do not have all of the properties that are typically ``taken for granted" with model based control laws. Closed loop dynamics obtained from PD control laws do not necessarily have equilibrium points. %; especially if the desired trajectories are time dependent. 
This also implies that orbits in the HZD may not necessarily be orbits in the full order dynamics. However, we can use properties of the inertia, Coriolis-centrifugal and gravity matrices, and then guarantee desirable convergence rates to an ultimate bound, i.e., exponential ultimate boundedness of the outputs. This property was utilized in \cite{whitcombpd1993} to establish boundedness of tracking errors for fully actuated systems.

\newsec{Organization.}
The paper is structured as follows. Section \ref{sec:hs} introduces the hybrid system model of locomotion and the associated control methodologies. This section also describes the concept of HZD and the associated stable periodic orbit. 
In Section \ref{sec:pd}, we focus on underactuation and construct the PD control law for output stabilization. We will also list a set of assumptions on the desired trajectories that will be useful to simplify the main results of the paper. Section \ref{sec:mainresults} contains the main results, i.e., stability results of PD based control laws for hybrid systems with Lagrangian dynamics in a series of Lemmas and Theorems. Proofs are provided in Section \ref{sec:proofs}. Finally Section \ref{sec:results} provides the simulation results of PD control on a $2$-DOF walker.

%%%%%%%%%%%%%%%%%%%%%%%%%%%%%%%%%%%%%%%%%%%%%%%%%%%%%%%%%%%%%%%%%%%%%%%%%%%%%%%%%%%%%%%%%%%%%%%%%
%% ROBOT MODEL SECTION
%%%%%%%%%%%%%%%%%%%%%%%%%%%%%%%%%%%%%%%%%%%%%%%%%%%%%%%%%%%%%%%%%%%%%%%%%%%%%%%%%%%%%%%%%%%%%%%%%
\section{Robot walking model and control}\label{sec:hs}

In this section, we will discuss the hybrid model of a walking robot, and the associated notion of hybrid zero dynamics (HZD) for walking. The associated periodic orbits of the HZD will also be described.

\newsec{Notation.}
 $\R$ is the set of real numbers, $\R^n$ denotes the Euclidean space of dimension $n$. 
 The open ball of radius $r>0$ centered at $x\in\R^n$ is denoted by $\B_r(x)$. Given $x\in\R^n$, $|x|$ is the Euclidean norm of $x$, and given a matrix $A\in\R^{n\times m}$, $\|A\|$ is the matrix norm of $A$.  Given a set $S\subset \R^n$, we denote the shortest distance between the point $x\in \R^n$ and the set $S$ to be
 $\|x\|_S : = \inf_{y\in S} \|x - y\|$. 
We will sometimes denote the vector $\begin{bmatrix} x^T, y^T \end{bmatrix}^T \in \R^{n_x + n_y}$ as the pair $(x,y)$. Note that the Euclidean norm has the property $|(x,y)|^2 = |x|^2 + |y|^2$. 

\subsection{Robot model}
We consider an $n$-DOF robotic system, with the configuration manifold $\Q$. 
We will specifically consider relative degree two systems for convenience. Therefore, we denote the state by $x=(q,\dq)\in T\Q$. We will denote the torque input by $u\in\R^m$, which is of dimension $m$. %In addition to the torque input, we also have $n_h$ holonomic constraint forces $\Lambda \in \R^{n_h}$ acting at various points on the robot (for example, foot contacts with ground act as holonomic constraints).
The dynamic model of walking consists of a continuous (swing) phase and a discrete (impact or foot-strike) phase. The discrete phase consists of a switch based on a guard condition (i.e., the swinging foot height $h$ crossing zero). Each of these events will be described briefly below.

\begin{figure}[!ht]
\centering
	\includegraphics[height=2.8cm]{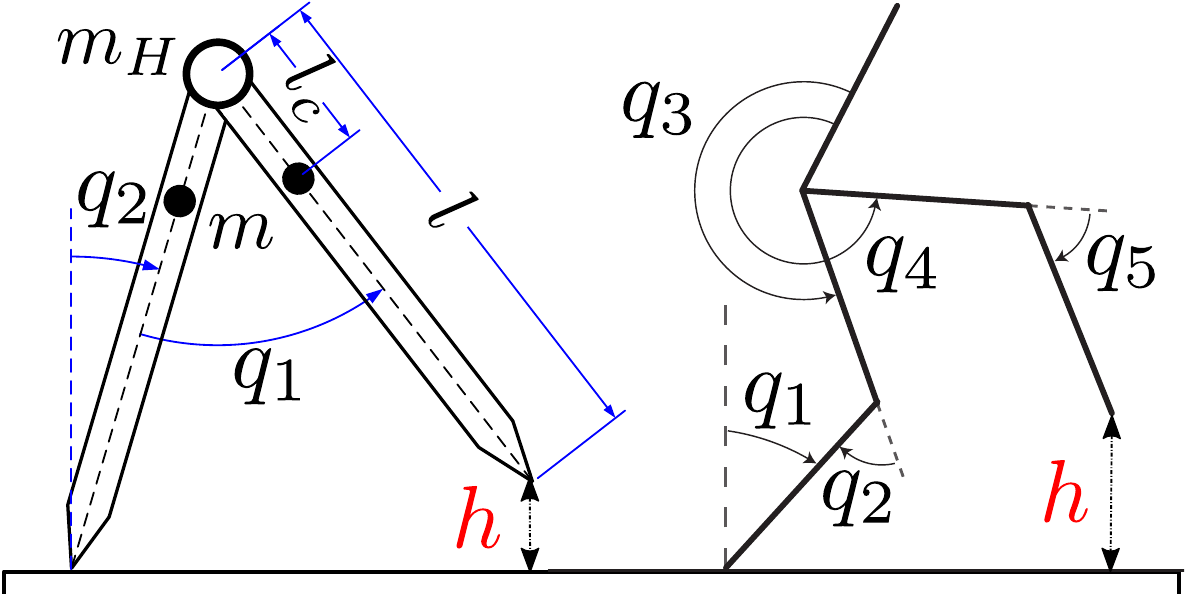}	
	\includegraphics[height=2.9cm]{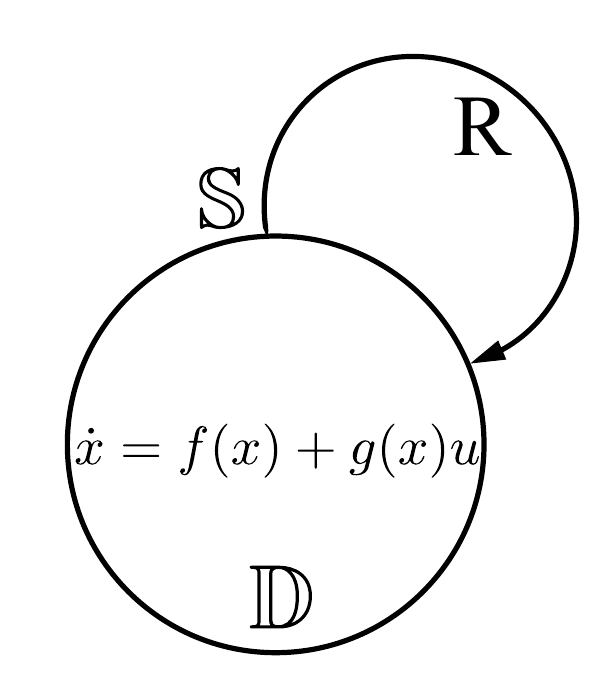}
	\caption{Figure showing a $2$-link (left) and a $5$-link walking robot (middle). One foot is in contact with ground while the other is at a height $h$ from the ground. A directed graph structure for a hybrid system model of walking is shown on the right. }
	\label{fig:amber1}
\end{figure}

\newsec{Continuous dynamics.} Given the states $(\q,\dq)$ and inputs $u$, the Euler-Lagrangian dynamics is given by
\begin{align}
  \label{eq:eom-general} 
  D(\q) \ddq + C(\q,\dq)\dq + G(\q) =  \Ba u,
\end{align}
%\todo[inline]{Think about including damping term $E$. AMBER1 uses P Voltage control and has motor model included.}
where $D(q) \in \R^{n\times n}$ is the inertia matrix, $C (\q,\dq) \in \R^{n\times n}$ is the Coriolis-centrifugal matrix, $G(q)\in \R^n$ is the gravity vector, and $\Ba \in \R^{n\times m}$ is the mapping of the torques to the joints. %, and $J_h(q)\in\R^{n_h\times n}$ is the Jacobian of the holonomic constraints. 
Without loss of generality, we assume that the choice of $\q$ is such that the mapping of torques to actuated joints is one-to-one i.e., each column of $\Ba$ consists of only one element with value one and the rest are zeros. Having described \eqref{eq:eom-general}, we have the following properties of the model \cite{ROB:ROB2,gravityboundedness,4339540}\footnote{The class of robots that satisfy these properties are described in \cite{ROB:ROB2,gravityboundedness}. %s always true for serial manipulators with pure revolute joints, pure prismatic joints, or
For example, this is true for serial manipulators with all of their prismatic joints preceding the revolute joints. Even for the prismatic joints, like in spring deflections, we know that these deflections are usually restricted by hardstops. This allows us to include a larger class of mechanical systems.}:
%%holonomic constraints by construction:
%\gap
\begin{property}\label{prop:dc}
	$D$ is positive definite symmetric, and $\dot D - 2C$ is skew-symmetric for any $(\q, \dq) \in T\Q$.
\end{property}
%\gap
\begin{property}\label{prop:1}
%For some $\underline{c}_d$, $\bar c_d $, $c_c> 0$, and $c_g > 0$
There exist positive constants $c_l$, $c_u > 0$ such that for any $(\q, \dq) \in T\Q$,
%\begin{align}
%\label{eq:constantscd}
%\underbar c_d \leq \|D(\q)\| \leq \bar c_d, \quad \|C(\q,\dq)\| \leq c_c |\dq|, \quad |G(\q)| \leq c_g.
\begin{itemize}
\item $c_l \leq \|D(\q)\| \leq  c_u$
\item $c_l \leq \|D^{-1}(\q)\| \leq  c_u$
\item $\| \dot D(\q) \| \leq c_u |\dq|$
\item $\|C(\q,\dq)\| \leq c_u |\dq|$
\item $|G(\q)| \leq c_u$.
\end{itemize}

\end{property}
%\gap
Note that each of the matrices, $D, D^{-1},C,G$ have their own bounds. We have used the same constants for ease of notations.
\eqref{eq:eom-general} can be represented in statespace form  as
\begin{align}
\label{eq:dynamicsmech}
%\left [ \begin{array}{c} \dq \\ \ddq \end{array} \right ] &= f (q,\dq) + g (q,\dq) u.
\dot x = f (x) + g (x) u,
\end{align}
by appropriate determination of $f,g$ (see \cite[(13)]{kolathaya2016parameter}).
%which is similar to \eqref{eq:system3}. Given the continuous dynamics of the robot, we 
The continuous dynamics of the robot is defined on the set of admissible states $\Domain\subset T\Q$, defined by
\begin{align}
    \Domain = \{ (\q,\dq) \in T\Q : h(\q,\dq) \geq 0 \},
\end{align}
where $h:T\Q \to \R$ is the height of the swing foot (see \figref{fig:amber1}). $\Domain$ is called the domain. Note that $h$ is chosen such that it only depends on the configuration $\q$ i.e., $\frac{\partial h}{\partial \dq}\equiv 0$\footnote{\addblue{In previous works, like in \cite{TAC:amesCLF}, $h$ is chosen such that $L_g h \equiv 0$. This includes a larger class of $h$, the analysis for which is beyond the scope of this paper.}}. %Note that $h$ can be any continuously differentiable function as long as $L_g h = 0$.

\newsec{Discrete dynamics.} Having described the continuous dynamics in statespace form \eqref{eq:dynamicsmech}, we can now describe the discrete dynamics of the walking robot. As observed in \figref{fig:amber1}, when the height $h$ crosses zero transversally, we have an impact. 
\addblue{This impact is represented with the rigid contact model of \cite{hurmuzlu:biped,hurmuzlu_marghitu}. The swinging foot is assumed to have no rebound or slip during an impact. The velocity component of the robot state experiences a jump, while the configuration component remains continuous. Since the walking is symmetric, the roles of the legs are swapped after every foot-strike.}
We define the guard set representing the foot-strike as
\begin{align}\label{eq:guard}
    \Guard = \{ (\q,\dq) \in \Domain : h(\q,\dq) = 0, L_f h(\q,\dq) <0 \},
\end{align}
with $L_g h(\q,\dq) \equiv 0$. Here $L_f, L_g$ are the Lie derivatives w.r.t. $f,g$ respectively. When $x\in\Guard$, we have a discrete event:
\begin{align}\label{eq:impactmap}
x^+ = {\rm{R}}(x), \quad x \in \Guard,
\end{align}
where $x$ is the pre-impact state, $x^+$ is the post-impact state, and ${\rm{R}}:\Guard \to \Domain$ is the impact map (or reset map) of the robot. 

Having obtained the continuous time \eqref{eq:dynamicsmech} and discrete time \eqref{eq:impactmap} model of walking, we obtain the hybrid control system model as
\begin{align}\label{eq:hs}
    \HybridControlSystem = \left \{ \begin{array}{cccc}
          \dot x &=& f(x) + g(x) u , &x\in\Domain \backslash \Guard \\
          x^+ &=& {\rm{R}}(x),  &x\in\Guard.
    \end{array} \right.
\end{align}
A pictorial representation of this hybrid control system model is given in \figref{fig:amber1}. Note that based on the definition of the guard set \eqref{eq:guard}, we assume that ${\rm{R}}(\Guard) \cap \Guard = \emptyset$ (to avoid consecutive jumps). %Given this hybrid system model, the goal is to realize a stable walking gait. %We will specifically focus on realizing a stable periodic orbit.

\subsection{Hybrid zero dynamics}

We will now mathematically describe the notion of HZD.
Denote the relative degree two outputs $y:\Q \to \R^m$ of dimension $m$ as
\begin{align}
	y(\q) = y^a (\q) - y^d(\q),
\end{align}
where $y^a:\Q \to \R^m$, $y^d: \Q \to \R^m$ are the actual and the desired values respectively. \addblue{The desired values are chosen such that $y(\q)=0$ for at least one $\q$, like in \cite[HH4]{WGK03}.} Similarly, denote the 
passive states (or the unactuated states) of dimension $2l = 2n - 2m$ as $z: T\Q \to \R^{2l}$. Accordingly, we have the output dynamics (or the transverse dynamics) and passive dynamics of the form:
\begin{align}\label{eq:outputdynamics}
	\ddot y = & L_f^2 y + L_g L_f y u \\
%\end{align}
%and the passive dynamics or the unactuated dynamics of the form:
%\begin{align}
\label{eq:passivedynamics}
	\dot z = & \psi(y,\dot y ,z).
\end{align}
Here $\psi:\R^{2n} \to \R^{2l}$ is the vector field (of $z$). \addblue{Note that $z$ is chosen such that $L_g z\equiv 0$ and a diffeomorphism from $(\q,\dq)$ to $(y,\dot y,z)$ exists (see Assumption \ref{ass:phidiff} further ahead). We denote this diffeomorphism as $\Phi: \Domain \to \R^{2n}$.}

With a suitable output stabilizing controller, we can guarantee convergence of the outputs $(y,\dot y)$ to zero. 
For example, a feedback linearizing control law of the form:
\begin{align}\label{eq:fblin}
	u_{\rm{IO}} = L_g L_f y^{-1} ( - L_f^2 y - 2 \epsilon \dot y - \epsilon^2 y),
\end{align}
yields exponential convergence of the outputs to zero.
% \begin{align}
% 	|(y(t),\dot y(t))| \leq \frac{1}{\epsilon}e^{-\epsilon t} |(y(0),\dot y(0))|,
% \end{align}
% where we have denoted $y(t)$ to be the flow of the resulting closed loop output dynamics.
% 
% Construction of the outputs $y$ is crucial for the realization of the resulting passive dynamics. In other words, 
If $(y,\dot y) = (0,0)$, the resulting passive dynamics given by
\begin{align}
\dot z  = \psi(0,0,z),
\end{align}
is now called the zero dynamics of the closed loop system of \eqref{eq:eom-general}. It is important to note that output stabilization can be realized only during the swing mode (continuous event) of the hybrid dynamics. On the other hand, if the outputs are chosen in such a way that the zero dynamics is invariant of the discrete dynamics, we have {\it hybrid zero dynamics} (HZD). Consider the reduced dimensional surface:
\begin{align}
\ZD{} = \{(\q,\dq) \in \Domain : y(\q)=0,  \dot y(\q,\dq) = 0  \}.
\end{align}
Also consider a post-impact map defined in terms of the transformed coordinates:
\begin{align}
 \ResetMap (y,\dot y,z) := \Phi \circ {\rm{R}} \circ \Phi^{-1} (y,\dot y,z),
\end{align}
which consists of two components:
\begin{align}
  \begin{bmatrix} \ResetMap_{y} (y,\dot y,z)\\ \ResetMap_{z} (y,\dot y,z)\end{bmatrix} := \ResetMap (y,\dot y,z),
\end{align}
corresponding to the outputs $(y,\dot y)$ and the zero coordinates $z$ respectively. We have HZD, if the following is satisfied:
\begin{align}\label{eq:HZDCond}
{\rm{R}}(\Guard \cap \ZD{}) \subset \ZD{} \quad {\rm{or}} \quad \ResetMap(\Phi(\Guard \cap \ZD{})) \subset \Phi(\ZD{}).
%\ResetMap_y(0,0,z^-) = (0,0), \forall z^-\in \Phi(\Guard \cap \ZD{}).
\end{align}
With this formulation, the goal now is to obtain a periodic orbit in the HZD and, consequently, a periodic orbit in the full order hybrid dynamics.
% we then have hybrid zero dynamics (HZD). $\ResetMap_z$ is obtained from \eqref{eq:impactmap} by expressing in terms of transformed coordinates $(y,\dot y, z) = \Phi(\q,\dq)$.

\subsection{Hybrid periodic orbits}
With the initial condition $(y^*,\dot y^*, z^*) \in \Phi(\Guard)$, and the control law \eqref{eq:fblin}, let $\phi_t(\ResetMap(y^*,\dot y^*, z^*))$ be the resulting flow of \eqref{eq:outputdynamics}, \eqref{eq:passivedynamics} represented in transformed coordinates. By assuming right continuity \cite[Section II-B]{veer2017poincare}, this flow is, in fact, the solution for the entire hybrid system \eqref{eq:hs} that includes both the continuous and the discrete dynamics. We say that there is a periodic orbit if there exists a $T^*>0$ such that %$\phi_{t}(\ResetMap(y^*,\dot y^*, z^*))$ exists for all $t\in [0,T^*)$ and
\begin{align}
\lim_{t\to T^*}\phi_{t}(\ResetMap(y^*,\dot y^*, z^*)) = (y^*,\dot y^*, z^*),
\end{align}
and the point $(y^*,\dot y^*, z^*)$ is called the fixed point of the periodic orbit. We will denote this periodic orbit as
\begin{align}
	\Orbit := \{ \phi_t(\ResetMap(y^*,\dot y^*, z^*)) \in \Phi (\Domain) :   0 \leq t < T^* \}.
\end{align}
We will also denote the fixed point in angle-velocity coordinates as $x^*:=\Phi^{-1}(y^*,\dot y^*, z^*)$. 

If the periodic orbit $\Orbit$ satisfies some properties (like transversality and isolated intersections with the guard $\Phi(\Guard)$ \cite[H2.4, H2.5]{MOGR05}, \cite[A.6, A.7]{veer2017poincare}), then we know that for an initial state $x_0$ in a small enough neighborhood of $x^*$ i.e., $x_0 \in \B_r(x^*)\cap \Guard$, the time-to-impact function is well defined with distinct lower and upper bounds, and can be obtained as
\begin{align}\label{eq:dwelltimeyydot}
T(y,\dot y,z) := \inf \{ t > 0 : h(\Phi^{-1}(\phi_t(\ResetMap(y,\dot y,z))) = 0 \},
\end{align}
where $(y,\dot y,z)= \Phi(x_0)$. See \cite{MOGR05,veer2017poincare} for more details on time-to-impact or dwell-time functions. Note that the problem formulation is constructed in such a way that undesirable behaviors like consecutive jumps and Zeno executions are avoided. Later on (Lemma \ref{lm:boundedness}, \ref{lm:zboundedness}), we will show that by using sufficiently large controller gains, even PD based control laws yield well defined time-to-impact functions.
%The neighborhood radius $r$ is very important to satisfy the conditions of implicit function theorem.

%Given the initial state $x_0 \in \B_r(x^*) \cap \Guard$, the goal is to obtain an output stabilizing controller that renders $\Orbit$ exponentially stable. 
Having defined the periodic orbit and time-to-impact functions, we can define some stability properties for $\Orbit$ that will be useful throughout the paper. For the following definition, we consider the solution $\phi_t$ for the entire time interval $[0,\infty)$ (by assuming right continuity \cite[Section II-B]{veer2017poincare}).
%\gap
\begin{definition}\label{def:eub}{\it 
	$\Orbit$ is said to be locally exponentially stable (LES) if there are constants $M, r,\lambda >0$ such that for all $(y,\dot y, z) \in \B_r (0,0,z^*) \cap \Phi(\Guard)$, %\footnote{Here $\|.\|_\Orbit$ is the nearest distance to $\Orbit$ defined as $\| a \|_\Orbit := \inf_{b\in \Orbit} \| a -b \|$. }
	\begin{align} 
	\|\phi_t(\ResetMap(y,\dot y, z))\|_\Orbit \leq M e^{-\lambda t} \|\ResetMap(y,\dot y, z)\|_\Orbit.
	\end{align}
	Similarly, $\Orbit$ is said to be locally exponentially ultimately bounded (LEUB) if there are constants $M, r,\lambda,  d >0$ such that for all $(y,\dot y, z) \in \B_r (0,0,z^*) \cap \Phi(\Guard)$,
	\begin{align} \label{eq:expobounded}
	\|\phi_t(\ResetMap(y,\dot y, z))\|_\Orbit \leq M e^{-\lambda t} \|\ResetMap(y,\dot y, z)\|_\Orbit +  d.
	\end{align}}
\end{definition}
%\gap

We will be mainly establishing LEUB of $\Orbit$ via \poincare maps (due to equivalence in stability results between periodic orbits and \poincare maps \cite{MOGR05,veer2017poincare}). Therefore, it is sufficient if we define the flow $\phi$ for the closed interval $[0,T]$, i.e., for only the continuous dynamics. After every impact, the time $t$ can be reset to zero with the resulting new initial state on the \poincare section. More details on \poincare maps are provided in Section \ref{sec:mainresults}.
%\gap
\begin{remark}\label{rm:leubremark}
{\it 
The notion of ultimate boundedness is valid even for systems without equilibrium points including periodic orbits \cite[4.8]{khalil2002nonlinear}. In the analysis that follows the next section, we will show that the set of points $\Orbit$ can be shown to be ultimately bounded when PD based control laws are applied.}
\end{remark}
%\gap

We can have similar notions of ultimate boundedness for the output coordinates in the continuous dynamics. 
For convenience, we will denote the initial state as $(y_0,\dot y_0, z_0)$ and the resulting trajectory, post-impact, as $(y(t), \dot y(t), z(t))$ with $(y(0),\dot y(0), z(0)) = \ResetMap(y_0,\dot y_0, z_0)$. We have the following definition for boundedness of the outputs:
%\gap
\begin{definition}\label{def:oeub}{\it
		Given $T_\delta >0$, the zero values of the outputs $(y(t),\dot y(t))$ are said to be locally exponentially ultimately bounded (LEUB) in the interval $t\in [0,T_\delta]$ if there are constants $M, r,\lambda,  d >0$ such that for all $(y_0,\dot y_0, z_0) \in \B_r (0,0,z^*) \cap \Phi(\Guard)$,
	\begin{align} \label{eq:outputbounded}
|(y(t),\dot y(t))| \leq M e^{-\lambda t} |(y(0),\dot y(0))| +  d.
\end{align}}
\end{definition}

The notion of ultimate boundedness (UB) and local exponential ultimate boundedness (LEUB) are applied to systems that are forward complete, i.e., $t \in [0,\infty)$. We would still like to use this definition for shorter finite intervals, since we have a hybrid system with restrictions on dwell-time. Therefore, whenever we say that the outputs are LEUB in the interval $[0,T_\delta]$, we mean that \eqref{eq:outputbounded} is valid for $[0,T_\delta]$.
%\gap

It was shown in \cite[Theorem 2]{TAC:amesCLF} that if there is an exponentially stable periodic orbit in the HZD, then by choosing a sufficiently large enough $\epsilon$ in \eqref{eq:fblin}, we can realize an exponentially stable periodic orbit in the full order dynamics. In this paper, we will particularly focus on obtaining a similar result (boundedness) with PD control by choosing large enough gains.

\section{PD tracking with underactuation}\label{sec:pd}

% For the case where $m < n$ (underactuation), we can pick the passive desired angle to be equal to the corresponding actual angle itself. Therefore, irrespective of the degree of actuation, we can obtain the same control formulation \eqref{eq:pdcontrollaw} with the desired configuration for the unactuated joints picked accordingly. 
% $q^u_d = q^u$.
In this section, we will focus on PD based control laws for robotic systems with underactuation. Since $m$ is the degree of actuation (DOA) and $n$ is the degree of freedom (DOF), we have the corresponding degree of underactuation as $l = n-m$. Accordingly, we have the following separation in the dynamics:
\begin{align}\label{eq:underactuatedeom}
D_{11}(\q) \ddot q^u + D_{12}(\q) \ddot q^a + C_1(\q,\dq) \dq + G_1(\q) &= 0 \nonumber \\
D_{21}(\q) \ddot q^u + D_{22}(\q) \ddot q^a + C_2(\q,\dq) \dq + G_2(\q) &= u,
\end{align}
where the terms corresponding to $D,C,G$ are apparent from the setup. $q^u$, $q^a$ are the unactuated and actuated configurations respectively. Also let $\Bu \in \R^{n\times l}$ be the constant matrix obtained in such a way that if the $l$ unactuated joints had inputs, then these inputs would be mapped via this matrix. \addblue{In other words, the new notations $B_c, \q^u, \q^a$ are defined in such a way that
\begin{align}
\q = \begin{bmatrix} B_c &  B \end{bmatrix} \begin{bmatrix}q^u \\ q^a \end{bmatrix}.
\end{align}}
Accordingly, we have that
\begin{align}
\begin{bmatrix}
D_{11} & D_{12} \\
D_{21} & D_{22} 
\end{bmatrix} & = \begin{bmatrix}
\Bu^T D \Bu  & \Bu^T D \Ba \\
\Ba^T D \Bu & \Ba^T D \Ba 
\end{bmatrix} \nonumber \\
\begin{bmatrix}
 C_1 \\
 C_2
\end{bmatrix} & = \begin{bmatrix}
		  \Bu^T C \\
		  \Ba^T C
		  \end{bmatrix} \nonumber \\
\begin{bmatrix}
 G_1 \\
 G_2
\end{bmatrix} & = \begin{bmatrix}
		  \Bu^T G \\
		  \Ba^T G
		  \end{bmatrix}.
\end{align}
The matrices $\Bu,B$ have some important properties that will be utilized in the proofs in this section. Some of them are
\begin{align}
 & [ \Bu, \Ba ] = \begin{bmatrix} \Bu^T \\ \Ba^T \end{bmatrix} =  \1_{n\times n} \nonumber \\
 & \Ba^T \Ba = \1_{m\times m}, \:\:\: \Bu^T \Bu = \1_{l \times l},  \:\:\: \Bu^T \Ba = \0_{l \times m}, \:\:\: \Ba^T \Bu = \0_{m \times l}, \nonumber
\end{align}
where $\1$ is the identity matrix, $\0$ is the zero matrix with appropriate dimensions. Based on these properties $\ddot q^u$ can be eliminated from \eqref{eq:underactuatedeom} to obtain
\begin{align}\label{eq:eliminateddynamics}
 \Ba^T A D\Ba \ddot q^a + \Ba^T A C \dq + \Ba^T A G =  u,
\end{align}
where
\begin{align}
 A := \1_{n\times n}  -  D  \Bu (\Bu^T D \Bu )^{-1} \Bu^T, \nonumber
% A(\q) := \1_{n\times n}  -  D (\q) \Bu (\Bu^T D(\q) \Bu )^{-1} \Bu^T, \nonumber
\end{align}
with $\1_{n\times n}$ being the identity matrix of dimension $n$. It can be verified that $\Ba^T AD\Ba$ is nothing but the Schur complement of $D_{11} = \Bu^T D \Bu$ in $D$, and it has some important properties:
%\gap
\begin{proposition}\label{prop:schurpd}
$B^T A(\q) D(\q) B$ is symmetric positive definite for all $(\q,\dq) \in T\Q$. 
\end{proposition}
%\gap
\begin{proposition}\label{prop:schurother}
%There exist positive constants $c_a, c_{ad}$ such that 
There exist positive constants $ c_l,c_u$ such that for all $(\q,\dq) \in T\Q$,
\begin{itemize}
  \item $ c_l \leq \| \Ba^T A(\q)D(\q) \Ba\| \leq {c}_{u}$
  \item $ c_l \leq \| (\Ba^T A(\q)D(\q) \Ba)^{-1}\| \leq {c}_{u}$
  \item $\|A(\q)\| \leq c_u$
  \item $\|A(\q)D(\q)\| \leq c_u$
  \item $\|\dot A(\q,\dq) D(\q) + A(\q) \dot D(\q,\dq)\|\leq c_{u}|\dq|$
  \item $\|\Ba^T A(\q) C(\q,\dq)\| \leq c_u  |\dq|$
  \item $\|\Ba^T A(\q) G(\q)\| \leq c_u $.
\end{itemize}
\end{proposition}
%\gap
\addblue{
See \cite[Theorem 2.1, Corollary 4.1]{LUinverseblockmatrix} and \cite[Appendix A.5.5]{boyd2004convex} for more details on Schur complement matrices. Note that, similar to Property \ref{prop:1}, we have used the same constants $ c_l, c_u$ for ease of notations. Proofs of Propositions \ref{prop:schurpd} and \ref{prop:schurother} are provided in Appendix \ref{FirstAppendix}.
}

%\gap

\subsection{Outputs and control}
Having represented the dynamics in the form \eqref{eq:eliminateddynamics} by eliminating $\ddot q^u$, we will now define the outputs for the robot. %The outputs correspond to the actuated degrees of freedom of the robot.
For the actuated configuration $q^a$, we define the following relative degree two outputs:
\begin{align}\label{eq:errordefined}
	e (\q^a,\tau) = \q^a - \q^a_d(\tau) ,
\end{align}
where $e$ defines the difference between the actual and the desired value of the actuated joints of the robot. The desired configuration $q^a_d : \R_{\geq 0} \to \R^m$ is a function of a variable called the phase (or the gait timing) variable $\tau:\Q \to \R$, which is a function of the configuration. %We will use the state dependent phase to obtain the HZD (explained more later). 
By a slight abuse of notations, we will sometimes remove the phase $\tau$ and denote the desired trajectories by $q^a_d(\q)$. %, which show the dependency on configuration. 

%\subsection{Control law}
Having defined the error $e$, we have its derivative as
\begin{align}
 \dot e(\q,\dq) = \dq^a  - \frac{\partial \q^a_d(\q)}{\partial \q} \dq = J(\q) \dq^a - \frac{\partial \q^a_d(\q)}{\partial \q^u} \dq^u. \nonumber
\end{align}
and $J(\q):=\frac{\partial e(\q)}{\partial \q^a}$ is the Jacobian w.r.t. $\q^a$. We use the following PD control law: 
\addblue{\begin{align}
\label{eq:pdtrackingunderactuated}
u_{\rm{PD}}(\q,\dq)  %&= -  K_p (q^a - q^a_d(\q)) - K_d (\dq^a - \dq^a_d(\q,\dq)) \nonumber \\
		      &  = - J(\q)^T K_p e(\q) - J(\q)^T K_d \dot e(\q,\dq),
\end{align}
where $K_p,K_d$ are the gain matrices of dimension $m$. 
\begin{remark}\label{rm:pdtracking}
 {\it If the desired configuration is only a function of $\q^u$, then $J(\q)=\1$, which reduces \eqref{eq:pdtrackingunderactuated} to the familiar form of PD control law. 
 }
\end{remark}}
For simplicity, we will assume that equal gains are applied for every joint i.e., $K_p = k_p \1$, $K_d = k_d \1$ for some $k_p$, $k_d>1$ and an identity matrix $\1$ of appropriate size. Having defined this PD control law \eqref{eq:pdtrackingunderactuated}, we have the resulting closed loop dynamics of \eqref{eq:hs} as
\begin{align}
\dot x  = f^{cl} (x) := f(x) + g(x)u_{\rm{PD}}(\q,\dq),
\end{align}
and with the initial state $x_0\in \Guard$,
we have the resulting flow (integral curve) of the continuous dynamics %$(q,\dq)$ as a function of time as 
as $x(t)=(q(t),\dq(t))$, where $(q(0),\dq(0))=R(x_0)$ and $t\in [0,T]$ with $T>0$ (see \eqref{eq:dwelltimeyydot}) being the time until the next impact.

\subsection{Hybrid zero dynamics}\label{subsec:zd}
%We will denote the zero coordinates to be $z$. 
For underactuated robotic systems, we typically choose the zero coordinates to be the following:
\begin{align}\label{eq:zerocoordinates}
	z(\q,\dq) := \begin{bmatrix}
	z_1(\q) \\ z_2(\q,\dq)
	\end{bmatrix} := \begin{bmatrix}
	q^u \\ \Bu^T D(\q) \dq
	\end{bmatrix}.
\end{align}
%We will denote the output coordinates to be 
It is important to note that the coordinates $z$ shown above are chosen based on $e$ (see Assumption \ref{ass:phidiff} ahead). Accordingly, we have the output zero coordinates as $(e,\dot e,z)$, and the corresponding transformation as $\Phi(\q,\dq)= (e,\dot e,z)$. With this transformation, we have the following passive dynamics:
\begin{align}\label{eq:zerodynamics}
\dot z  & = \psi(\Phi(\q,\dq)), \quad & (\q,\dq) & \in \Domain \backslash \Guard \nonumber \\
z^+ & = \ResetMap_{z} (\Phi (\q, \dq)), \quad & (\q,\dq) & \in \Guard,
\end{align}
where $\psi$ is 
% obtained from \eqref{eq:passivedynamics} and is 
given by
\begin{align}\label{eq:psidynamics}
\psi(\Phi(\q,\dq)) :=	\begin{bmatrix} \dot \q^u \\
	\Bu^T \left ( \dot D(\q,\dq) \dq  -  C(\q,\dq) \dq - G(\q) \right ) 
	\end{bmatrix}.
\end{align}
Here the input $u$ will not appear due to underactuation. If the outputs are zero i.e., $(e,\dot e) = (0,0)$, and if the hybrid invariance conditions \eqref{eq:HZDCond} are satisfied (with the notations $y, \dot y$ replaced with $e, \dot e$), we have hybrid zero dynamics (HZD). %If $\tau$ is state based, then we have {\it state based} HZD, and if it is purely time based, we have {\it time based} HZD.
The HZD obtained lies on the reduced dimensional surface:
\begin{align}
    \ZD{} = \{ (\q,\dq) \in \Domain : e(\q) = 0, \dot e(\q,\dq) = 0 \}.
\end{align}

Stability of HZD has been extensively studied, where the desired values $\q^a_d$, and the phase $\tau(\q)$ are chosen in such a way that the HZD has an exponentially stable periodic orbit \cite{grizzle20103d,kaveh_expo_ijrr,WGK03}.
Therefore, given that there is a constructive way to realize an exponentially stable periodic orbit in the HZD, we make the following assumption:
\begin{assumption}\label{ass:hzdstatebased}{\it
	The hybrid zero dynamics given by
	\begin{align}\label{eq:hzds}
	\mathcal{Z} := \left \{\begin{array}{ccccc}
	\dot z  &=& \psi(0,0,z), \quad  & (0,0,z)   \in \Phi(\ZD{} \backslash (\ZD{}\cap \Guard)) & \\
		z^+ &=& \Delta_z (0,0,z), \quad & (0,0,z)  \in \Phi(\ZD{} \cap \Guard) &	\end{array} \right. ,
	\end{align}
	has an exponentially stable periodic orbit, $\Orbit_z$, transverse to $\Phi(\ZD{}\cap \Guard)$. }
\end{assumption}
%\gap
\begin{remark}{\it 
Periodic orbits on the HZD are usually obtained through an offline optimization problem \cite{Hereid_etal_2016}. Since the model is not accurately known, Assumption \ref{ass:hzdstatebased} may seem restrictive. Relaxation of the above assumption with uncertainties incorporated have been studied in \cite{kolathaya2015parameter}, where the periodic orbit was shown to be exponentially ultimately bounded. In order to keep focus on the stability analysis for PD control laws, we will continue to use Assumption \ref{ass:hzdstatebased} for the rest of the paper.}
\end{remark}
%\gap

Denote the integral curve of the continuous zero dynamics as $\phi_t^{\rm{HZD}} : \R^{2l} \to \R^{2l}$. Since the HZD has a periodic orbit, we have a fixed point $z^*$, and the associated period $T^*$ that satisfies: $\phi^{\rm{HZD}}_{T^*}(\ResetMap_z(0,0,z^*)) = z^*$. We have this periodic orbit defined as
\begin{align}\label{eq:periodicorbitreduced}
\Orbit_z = \{\phi^{\rm{HZD}}_t(\ResetMap_z(0,0,z^*))  \in \R^{2l} : 0 \leq t < T^*  \}.
\end{align}

\subsection{Exponential stability via \poincare maps} 
Exponential stability of periodic orbits is characterized by using \poincare maps \cite{MOGR05,veer2017poincare}. Therefore,
%given the flow $\phi^{\rm{HZD}}_t(\ResetMap_z(0,0,z))$, we can ob
we define the following \poincare map:  
%We can also define the reduced \poincare map: $$\rho: \R^{2l} \to \R^{2l},$$ defined from the flow on the HZD \eqref{eq:hzds}:
\begin{align}\label{eq:pcarehzd}
\rho(z_s) := \phi^{\rm{HZD}}_{T_\rho(z_s)}(\ResetMap_{z}(0,0,z_s)).
\end{align}
% defined from the flow $\phi^{\rm{HZD}}_t(\ResetMap_z(0,0,z_s))$. 
Here $(0,0,z_s)\in \Phi(\ZD{} \cap \Guard)$ is the initial state, and $T_\rho$ is the reduced time-to-impact function given by
\begin{align}
T_\rho(z_s) = \inf\{ t >0:  h(\Phi^{-1} (\iota ( \phi^{\rm{HZD}}_t(\ResetMap_{z}(0,0,z_s))))) =0 \},
\end{align}
obtained similar to \eqref{eq:dwelltimeyydot}. Note that we have used the subscript $s$ in $z_s$ to distinguish it from the evolution of $z$ of the full order dynamics.
Since $\Orbit_z$ is exponentially stable, by the converse Lyapunov theorem for discrete systems, there exists an exponentially convergent Lyapunov function $V_z$ for some $r > 0$, % (possibly smaller than previously determined),
and positive constants $c_1,c_2,c_3,c_4$ such that for all $z_s \in \B_r(z^*) \cap (\ZD{} \cap \Guard)$,
\begin{align}
\label{eq:discreteVrho}
&    c_1 |z_s - z^* |^2 \leq V_z(z_s - z^*)  \leq  c_2 |z_s - z^* |^2 \nonumber \\
&    V_z (\rho(z_s) - z^*) - V_z(z_s - z^*) \leq  -c_3 |z_s - z^*|^2  \\
&    | V_z (z_s - z^*) - V_z (z'_s - z^*) | \leq  c_4 |z_s -z'|.(|z_s - z^*| + |z'_s - z^*|).\nonumber
\end{align}
In addition, we know that there exist constants $c_5>0$, $\gamma \in (0,1)$ such that
\begin{align}\label{eq:rhodynamics}
|\rho^i(z_s) - z^*| \leq c_5 \gamma^i |z_s- z^*|.
\end{align}
We will be mainly using this property along with the \poincare map of the full order hybrid system (described in the next section) to establish our main results. 

\subsection{Hybrid dynamics}
We can reconstruct $\Orbit$ from $\Orbit_z$ by the canonical embedding $\iota(z) := (0,0,z)$, i.e.,  $\Orbit = \iota (\Orbit_z)$. Since $\Orbit$ is not necessarily an orbit of the full order hybrid system, we will denote the point: $(0,0,z^*)$, simply as a nominal point of $\Orbit$.
%This also implies that the fixed point of $\Orbit$ is $(0,0,z^*)$. 
% where $\ZD{} = \{ (\q,\dq) \in \Domain : e(q^a,\tau(\q)) = \dot e(\dq^a,\tau(\q),\dot \tau(\q)) = 0 \}$ is the reduced dimensional surface obtained when the state based outputs are zero.
% 
%\todo[inline]{Add definition for Phase to state stability and show that the time based HZD is e-PSS given that state based HZD is ES.}
Given that $\mathcal{O}_z$ is exponentially stable, we are interested in the stability properties of $\Orbit$ when a PD control law of the form \eqref{eq:pdtrackingunderactuated} is applied. Accordingly, we have the following hybrid dynamics expressed in terms of the transformed coordinates $(e,\dot e,z)$:
\begin{align}\label{eq:hsoutputzero}
\HybridSystem = \left \{ \begin{array}{cl}
\ddot e &= f_e (e,\dot e,z) \nonumber \\
 & \hspace{10mm} + g_e(e,\dot e,z) u_{\rm{PD}}(\Phi^{-1}(e,\dot e,z))   \\
\dot z & =  \psi(e,\dot e,z) , \qquad   \\
 & \hspace{14mm} {\rm{when}}\hspace{1mm} (e,\dot e, z)\in \Phi(\Domain \backslash \Guard) \\
(e^+ , \dot e^+ , z^+)
&= \ResetMap(e,\dot e, z),   \\
 & \hspace{14mm} {\rm{when}} \hspace{1mm} (e,\dot e, z)\in \Phi(\Guard)
\end{array} \right. , \\
\end{align}
where $f_e, g_e$ are obtained from the Lie derivatives of $e$. For the continuous dynamics in \eqref{eq:hsoutputzero}, we will use $e(t), \dot e(t), z(t)$ in place of the flow $\phi$ for convenience.
The value at $t=0$, i.e., $(e(0),\dot e(0),z(0))$, are the post-impact states of the initial state $(e_0,\dot e_0, z_0) := \Phi(x_0)$. We will also denote the trajectory on the orbit $\Orbit$ by $(\q^*(t), \dq^*(t))$, and in the transformed coordinates by $(0,0,z^*(t))$. The following additional assumptions are used to establish the main result. 

\subsection{Additional assumptions on $\Orbit_z$ and the outputs $e$}
% Note that the following assumptions are valid for one step only i.e., time $t$ is reset to $0$ after every foot strike.
% %\gap
% \begin{assumption}\label{ass:timebasedphase} {\it The phase $\tau(\q)$ 
% 		is bounded for a local set of $\q, \dq$.}
% %		\item There exists $\tau^*>0$ such that  $\tau^t(t), \dot \tau^t(t)\leq \tau^*$ for all $t$
% 		%\item $\tau^t(t)$ is reset to $0$ after every guard-strike event.
% 		%		\item The time based desired angles and velocities are upper bounded.
% %	\end{itemize}}
% \end{assumption}
%\gap
% Assumption \ref{ass:timebasedphase} will be useful to establish the relationship between the time and state based outputs $e, e_t, \dot e, \dot e_t$. We will let $\tau^*>0$ be the bounds for both $\tau^t(t), \dot \tau^t(t)$ for all $t$\footnote{It is also sufficient if we have $\tau^t,\dot \tau^t$ bounded in the interval $[0,T_{\max}]$ (described more later), due to the fact that the time range in each step is upper bounded by $T_{\max}$. Practical implementations of PD based control laws use saturations if the time exceeds $T_{\max}$.}. % \in [0,T]$. 
% In addition, $\tau^t(0)=0$, which ensures that the desired trajectories are reset after every step. 
% With this assumption, we also 
\addblue{
We make the following assumptions on $e$. In particular, these assumptions are for a local configuration space of $\q$, i.e., we will pick a tube of radius $r$ around the orbit $\q^*(t)$:
\begin{align}
\mathbb{N}_r  = \{ \q  \in \Q : \inf_{t \in [0,T]} \| q - q^*(t) \| \leq r \}.
\end{align}
% These assumptions are for a local configuration space of $\q$, i.e., we will pick a tube of radius $r$ around the projection of $\Orbit$ to the configuration space:
% \begin{align}
% \mathbb{N}_\q  := \{ \q  \in \Q : \| \Phi(\q,0)\|_\Orbit \leq r \}.
% \end{align}
% \begin{assumption}\label{ass:boundeddesired}{\it 
% 		The desired actuated configuration $q^a_d(\q)$ is bounded by some $c_q>0$ for all $\q \in \mathbb{N}$. %\mathbb{N}_\q$. %In addition, the periodic flow $z^*(t):=\phi^{\rm{HZD}}_t(\ResetMap_z(0,0,z^*))$ with period $T^*$ on the orbit $\Orbit_z$ is also bounded by $k_q>0$ for all $t\in [0,T^*]$.
% 	}
% \end{assumption}
% This assumption will be useful to establish Proposition \ref{proposition:dece} and \ref{prop:schurpde} ahead.
% This is not a restrictive assumption and we typically impose this by saturating the phase $\tau(\q)$. This is well practiced in experiments (see \cite{hscc17running}, for example). 
We have the following first assumption on the outputs $e$ \eqref{eq:errordefined}:
\begin{assumption}\label{ass:phidiff}
 {\it
     The transformation $(\q^u, e(\q)):\Q \to \R^n$ is a local diffeomorphism onto its image for every $\q \in \mathbb{N}_r$. %In addition, the inverse of this diffeomorphism $\Phi^{-1}(q^u,e)$ is globally Lipschitz w.r.t. $q^u,e$.
 }
\end{assumption}
By Assumption \ref{ass:phidiff}, the Jacobian w.r.t. $\q$ is nonsingular:
\begin{align}
 J_e(\q) := %\frac{\partial \phi(\q)}{\partial \q} = 
 \begin{bmatrix}
                                      \1_{l \times l} & \0_{l \times m} \\
                                      \frac{\partial e(\q)}{\partial \q^u} & J(\q)
                                     \end{bmatrix}.
\end{align}
As a result, the continuous dynamics \eqref{eq:eom-general} can be reformulated and represented in terms  of $\q^u, e$ as
\begin{align}\label{eq:eom-general-in-terms-of-error-outputs}
 D_{e}(\q) \begin{bmatrix}
        \ddot \q^u \\ \ddot e
       \end{bmatrix}
 + C_{e}(\q,\dq) \begin{bmatrix}
        \dot \q^u \\ \dot e
       \end{bmatrix} + G_{e}(\q) = J^{-T}_e(\q) B u,
\end{align}
where  $ D_{e}=  J_e^{-T} D J_e^{-1}$, $C_{e} = J_e^{-T} C J_e^{-1} + J_e^{-T} D \frac{ d (J_e^{-1})}{dt}$, $ G_{e} = J_e^{-T} G$. As a result of this change of variables, the following functions will appear frequently throughout the paper:
\begin{align}\label{eq:bebeprime}
A_e (\q) &:= \1_{n\times n}  -  D_e(\q)  \Bu (\Bu^T D_e(\q) \Bu )^{-1} \Bu^T \nonumber \\
 B_e(\q) &:= \1_{m\times m} + \Ba^T D_e(\q)  \Bu (\Bu^T D_e(\q) \Bu )^{-1} \frac{\partial e(\q)}{\partial \q^u}^T  \\
 B_e'(\q) &:=	\1_{l \times l} - (\Bu^T D(\q) \Bu)^{-1} \Bu^T D(\q) \Ba J^{-1}(\q) \frac{\partial e(\q)}{\partial \q^u}.\nonumber
  \end{align}
Similar to \eqref{eq:eliminateddynamics}, we can eliminate $\ddq^u$ from \eqref{eq:eom-general-in-terms-of-error-outputs} to obtain
\begin{align}\label{eq:eliminatederrordynamics}
 \Ba^T A_e D_e \Ba \ddot e + \Ba^T A_e C_e \begin{bmatrix} \dot \q^u \\ \dot e\end{bmatrix} + \Ba^T A_e G_e = B_e J^{-T} u,
\end{align}
where $A_e, B_e$ are obtained from \eqref{eq:bebeprime}.  
Having defined $A_e, B_e$, we now impose the following assumption: % on $B_e$:
\begin{assumption}\label{ass:by}
{\it The outputs $e$ are chosen such that the norm $\| \frac{\partial e(\q)}{\partial \q^u}\|$ is sufficiently small. In particular, this norm is chosen such that for all $(\q,\dq)\in \bigcup_{\q\in\mathbb{N}_r} T_\q \Q$, 
	\begin{itemize}
	\item[1.] $B_e'$ is invertible, and
%	\item[2.] $B_e(\q) + B_e^T(\q)$ is positive definite.
	\item[2.] $ \Lambda_{B_e} := \begin{bmatrix}
				B_e + B_e^T & B_e + (1 + |e|)(B_e^T - \1) \\
				B_e^T + (1+|e|)(B_e - \1) & (1 + |e|)(B_e + B_e^T)
				\end{bmatrix} $ is symmetric positive definite.
%	\item[3.] $ \lambda_{\rm{ratio}}(\q) < 1$
	\end{itemize}
where $B_e',B_e$ are dependent on $\q$ (given by \eqref{eq:bebeprime}). %, and $$\lambda_{\rm{ratio}}(\q) := \frac{\|B_e(\q) - \1_{m \times m}\| \|B_e(\q)\|}{\lambda_{\min}((B_e(\q)+B_e^T(\q))/2)^2}.$$
}
\end{assumption}
Assumption \ref{ass:by} only requires that $e$ has small variations w.r.t. $\q^u$. This will be verified with the example biped model in Section \ref{sec:results}. We choose $e$ such that the entries of $B_e-\1_{m \times m}, B_e'-\1_{l \times l}$ are small in magnitude. By Gershgorin's circle theorem \cite[Theorem 2.1]{brakken2007gershgorin},  the diagonal entries of $B_e',B_e$ are positive and dominate the off-diagonal entries, thereby rendering them invertible. Furthermore, since $\Lambda_{B_e}$ is symmetric, the eigenvalues are real and positive, thereby ensuring positive definiteness. Note that this assumption is sufficient but not necessary, and its relaxed formulations are possible, which will be a subject of future work.}

\addblue{
Since the norm is a continuous function, $\|J_e(\q)\|$ has nonzero upper and lower bounds on a compact set $\mathbb{N}_r$. We can make use of this property to establish the following:
% Differentiating $e$ with respect to time gives 
% \begin{align}
%     % \ddot e(t,\q,\dq) &=  J_e(t,\q) \ddq + \dot J_e(t,\q) \dq -\frac{\partial }{\partial \q} \left (\frac{\partial q_d(t,\q)}{\partial t}\right) +  \frac{\partial^2 q_d(t,\q)}{\partial t^2},
% \ddot e =  J(\q) \ddq + \dot J(\q,\dq) \dq,
% \end{align}
% By Assumption \ref{ass:jacbounded}, we have that $\|J_e\|$ is bounded from both above and below in the space $\mathbb{N}_\q$. 
% The matrices $D_{e}$, $C_{e}$, $G_e$ have similar properties as that of $D,C,G$: 
\begin{proposition}\label{proposition:dece}
	{\it
		%Consider the error defined by \eqref{eq:configerror} satisfying assumptions F.1-F.6, with the resulting error dynamics given by \eqref{eq:eom-general-in-terms-of-error-outputs}. 
		$D_e$ is symmetric positive definite, and $\dot D_e  - 2 C_e$ is skew-symmetric. In addition, there exist $c_l, c_u >0$ (possibly smaller, larger than previously determined $c_l, c_u$) such that for any $(\q,\dq)\in \bigcup_{\q\in \mathbb{N}_r} T_\q \Q$,
		\begin{itemize}
			\item[1.] $c_l \leq \|D_e(\q)\| \leq  c_u$
			\item[2.] $c_l \leq \|D^{-1}_e(\q)\| \leq  c_u$
			\item[3.] $\| \dot D_e(\q) \| \leq c_u (|\dq^u| +  |\dot e|)$
			\item[4.] $\|C_e(\q,\dq)\| \leq c_u(|\dq^u| +  |\dot e|)$
			\item[5.] $|G_e(\q)| \leq c_u$.
		\end{itemize}
	}
\end{proposition}
Note that we have not restricted the tangent space $T_\q \Q$ to a compact set. This will be useful for allowing larger velocity variations (see Remark \ref{rm:epsiloninfty} ahead). Proof of Proposition \ref{proposition:dece} is provided in Appendix \ref{SecondAppendix}. The interested reader may also see \cite[Chapter 4, Section 5.4]{MLS94} for more details. }

\addblue{Similar to Proposition \ref{proposition:dece}, we have the following properties for $\Ba^T A_e D_e \Ba$:
\begin{proposition}\label{prop:schurpde}
	{\it
		 $\Ba^T A_e D_e \Ba$ is  symmetric positive definite. In addition, there exist positive constants $ c_l,c_u>0$ such that for all $(\q,\dq) \in \bigcup_{\q\in\mathbb{N}} T_\q \Q$, % $D_y, C_y, G_y$ satisfy %Property \ref{prop:1}.1-\ref{prop:1}.5.
		\begin{itemize}
			\item $ c_l \leq \| \Ba^T A_e(\q) D_e(\q) \Ba\| \leq {c}_{u}$
			\item $ c_l \leq \| {(\Ba^T A_e(\q) D_e(\q) \Ba)}^{-1}\| \leq {c}_{u}$
			\item $\|  \Ba^T \dot A_e(\q,\dq) D_e(\q) \Ba + \Ba^T  A_e(\q) \dot D_e(\q,\dq) \Ba \| \leq c_{u}(|\dq^u| + |\dot e|)$
			\item $\|\Ba^T A_e(\q) C_e(\q,\dq)\| \leq c_u  (|\dq^u| + |\dot e|)$
			\item $\|\Ba^T A_e(\q)G_e(\q)\| \leq c_u. $
		\end{itemize}
	}
\end{proposition}
Proof is provided in Appendix \ref{ThirdAppendix}.
With these propositions, we are now ready to present the main results of the paper.}

\section{Main results}\label{sec:mainresults}
The main results will be in a series of Lemmas and Theorems. %We will first revisit the notations used so that the main results presented are clear.
%%\gap
%%\gap
Since we are interested in a local result, we start from a small neighborhood of $(0,0,z^*)$ i.e., $(e_0,\dot e_0,z_0)\in \B_r(0,0,z^*) \cap \Phi(\Guard)$.
% , where
% \begin{align}\label{eq:bstar}
% %\B_*(r,\epsilon) := \{ (e, \dot e,z) \in \Phi(\Guard) : \left |\begin{bmatrix}\epsilon e \\ \dot e \\ \epsilon (z_1-z^*_1) \\ z_2 - z^*_2 \end{bmatrix} \right | < r   \},
% \B_r := \{ (e, \dot e,z) \in \Phi(\Guard) : |(e, \dot e, z- z^*)|< r   \}.
% \end{align}
The neighborhood radius $r$ may, perhaps, be smaller than previously determined.
Later on, $r$ may be reduced further depending upon the gains $k_p, k_d$.
%where $\epsilon >1$ is a user-definable constant. $\epsilon$ will also be used to substitute the gains $K_p, K_d$ later on. This formulation allows us pick the initial states only from specific types of sets, which are like the Lyapunov sublevel sets. We will describe the reasons for using this type of sublevel sets in Remark \ref{rm:repsilon}. 
With this initial condition, we have the following time-to-impact (dwell-time) function:
\begin{align}\label{eq:dwelltime}
	T(e_0,\dot e_0,z_0) :=\left \{ \!\!\! \begin{array}{ll}
\inf \{ t > 0 :  \!\!\! &h(\Phi^{-1}(e(t),\dot e(t),z(t))) = 0 \},  \\
& {\rm{if}} \: \exists \: t \: {\rm{s.t.}} \: (e(t),\dot e(t),z(t))\in \Phi(\Guard) \\
\infty , & {\rm{otherwise,}}
	\end{array}  \right.
\end{align}
where $(e(0),\dot e(0),z(0)) = \ResetMap(e_0,\dot e_0,z_0)$. %The height $h$ only depends on the configuration $\q$, and consequently only on $z_1(t), e(t)$. In other words, $h$ does not depend on the velocity output $\dot e(t)$, which will be useful in proof of Lemma \ref{lm:zboundedness} that follows. %$T$ is well-defined via implicit function theorem \cite[Proposition 2]{veer2017poincare} if we pick a small enough $r$. In addition, $T$ can be bounded by some constants $T_{\min}, T_{\max} >0$ such that $T_{\min} \leq T \leq T_{\max}$. 
Since ${\rm{R}}(\Guard) \cap \Guard = \emptyset$, we know that there is a $T_{\min}>0$ such that $T \geq T_{\min}$ (later we will also show that upper bound for $T$ also exists). 
We have the following first result. We show here that the outputs are LEUB (Definition \ref{def:oeub}).
%\gap
\begin{lemma}\label{lm:boundedness}{\it
%	Let $\Orbit_z$ be an exponentially stable periodic orbit in the hybrid zero dynamics \eqref{eq:hzds} transverse to $\Guard \cap \ZD{}$. 
%	In addition, 
Let the system \eqref{eq:hsoutputzero} be given, and let the desired configuration $q^a_d$ be chosen such that Assumptions \ref{ass:phidiff}-\ref{ass:by} are satisfied. Then there 
%	exist constants $k_e, k_s >0$, that are independent of the 
exist sufficiently large enough gains $k_p, k_d>1$, and a correspondingly small enough $r>0$, $T_\delta >0$, %\cup \{\infty\}$ 
such that for all	$(e_0,\dot e_0,z_0)\in \B_r(0,0,z^*) \cap \Phi(\Guard)$, %$(e,\dot e,z)\in \B_r$ \eqref{eq:bstar}, % \B_r(0,0,z^*)\cap \Phi(\Guard)$
%and for all $ \in \B_r(x^*) \cap \Guard$:
% \begin{itemize}
% 	\item[1.] The time based outputs $(e_t(t), \dot e_t(t))$ are LEUB in the interval $t \in [0,T_{\delta}]$.
% 	\item[2.] The state based outputs $(e(t), \dot e(t))$ are LEUB in the interval $t \in [0,T_{\delta}]$.
% \end{itemize} 
the outputs $(e(t), \dot e(t))$ are LEUB in the interval $t \in [0,T_{\delta}]$.}
%resulting solution of the closed loop dynamics of \eqref{eq:hsoutputzero} satisfies the following for all 
%
%	\begin{align}
%   	|(e_t(t),\dot e_t(t))| & \leq e^{- }, \nonumber \\
%%   	|(e(t),\dot e(t))| & \leq k_e, \nonumber \\
%%		|z(t)- \phi^{\rm{HZD}}_t(\ResetMap_z(0,0,z^*))| & \leq k_e.
%	\end{align}}
%In addition, $k_e,k_s$ are independent of the choice of the gains $k_p,k_d$.
\end{lemma}
Note that \cite[3.2]{khalil2002nonlinear} establishes boundedness by Gronwall-Bellman result for any finite interval (any arbitrary $T_\delta$) by choosing an initial point $(e_0, \dot e_0, z_0)$ very close to the nominal point $(0,0,z^*)$ (i.e., by reducing $r$). We cannot use this approach due to the fact that $\Orbit$ is not necessarily an orbit for \eqref{eq:hsoutputzero}. %This lemma provides an upper bounded on the time and state based states $e, e_t$ and also the zero coordinate $z$ in the interval $[0,T_{\max}]$. In addition, 
%Since the orbit $\Orbit$ is a bounded set we can obtain a constant $k_s>0$ such that $|(\q(t), \dq(t))| \leq k_s$ for all $t\in [0,T_{\max}]$. 
On the other hand, the following Lemma will, in fact, allow us to simply pick larger $k_p, k_d$ for stretching $T_\delta$ to $T$, %that ensure the resulting solution is as close to $\Orbit$ as possible.
%pick any arbitrary $T_\delta > 0$, 
thereby establishing ultimate boundedness of the outputs for the entire step.

\begin{lemma}\label{lm:zboundedness}{\it 
%Given that the initial state $x_0$ starts in the neighborhood $\B_r(x^*) \cap \Guard$, if the resulting solution of the state based outputs $(e(t), \dot e(t))$ are exponentially ultimately bounded in the interval $[0,T_\delta]$, then 
Let the system \eqref{eq:hsoutputzero} be given, and let the desired configuration $q^a_d$ be chosen such that Assumptions \ref{ass:phidiff}-\ref{ass:by} are satisfied. 
Given that $\Orbit_z$ is an orbit of the HZD, and $k_p, k_d, r, T_\delta>0$ are chosen such that the outputs $(e(t), \dot e(t))$ are LEUB in the interval $[0,T_\delta]$, we then have the following: 
\begin{itemize}
\item[1.] If $(e_0, \dot e_0, z_0) = (0,0,z^*)$, then there exist $C_k, C_t>0$ with $C_k$ dependent on $k_p, k_d$ such that %sufficiently large enough gains $K_p, K_d$ such that
\begin{align}\label{eq:zerrordynamics}
|z(t) - z^*(t)| \leq C_k(k_p, k_d) e^{C_t t} T_\delta, \quad t\in [0,T_\delta],
\end{align}
where $C_k$ decreases with increasing $k_p, k_d$.
\item[2.] For every $\delta>0$, there exist %$\epsilon>1$ and 
$k_p, k_d>1$, greater than previously determined, such that 
\begin{align}\label{eq:tstartdelta}
| T(0,0,z^*) - T^* | < \delta.
\end{align}
\item[3.] There exist $k_p, k_d>1$, greater than previously determined, and $r>0$, such that for all $(e_0,\dot e_0,z_0)\in \B_r(0,0,z^*) \cap \Phi(\Guard)$, %$(e, \dot e, z) \in \B_r$,
%For any finite $T$, there exists a sufficiently small $r$ and sufficiently large gains $K_p,K_d$ such that 
%$T(e, \dot e, z)<\infty$ \eqref{eq:dwelltime}, and
%$T_\delta = T(e, \dot e, z)$, i.e., 
%Lemma \ref{lm:boundedness} 
%and \eqref{eq:zerrordynamics} are
%is valid for all $t\in [0,T]$.
\begin{align}\label{eq:timeineq}
T_{\min} \leq T(e_0, \dot e_0 , z_0) \leq T_{\max},
\end{align}
for some pre-defined constants $T_{\max} > T^* > T_{\min}  > 0$.
\end{itemize}}
\end{lemma}
Lemmas \ref{lm:boundedness} and \ref{lm:zboundedness} together show that %by varying $r,K_p,K_d$, we can get the solution 
$(e(t), \dot e(t), z(t))$ can be close to $\Orbit$ for the entire step. %With this result, it is straightforward to prove the following Lemma:
%\gap
%\begin{lemma}\label{lm:timelemma}{\it 
%Let the system \eqref{eq:hsoutputzero} be given with Assumption \ref{ass:hzdstatebased}, in addition to $q^a_d, \tau^s, \tau^t$ satisfying Assumptions \ref{ass:timebasedphase}-\ref{ass:phasematch}. Then there exist $r> 0$, $ \epsilon >1$, and sufficiently large gains $K_p, K_d$ such that for all $(e, \dot e, z) \in \B_r$,
%%Given that the trajectory $(e(t), \dot e(t), z(t))$ can be brought arbitrarily close to $\Orbit$ by varying $r,K_p, K_d$, the time-to-impact function \eqref{eq:dwelltime} is well defined for all $(e, \dot e, z) \in \B_*(r,\epsilon)$ with well defined upper and lower bounds $T_{\min}, T_{\max} >0$.
%\begin{align}
% T_{\min} \leq T(e, \dot e , z) \leq T_{\max},
%\end{align}
%for some constants $T_{\max} > T^* > T_{\min}  > 0$.
%}
%\end{lemma}
%\gap
%
%Both the Lemmas \ref{lm:boundedness}-\ref{lm:zboundedness} focus on obtaining the neighborhood and the PD gains. % such that the resulting trajectory is arbitrarily close to $\Orbit$ in the continuous dynamics. 
The ensuing lemma and the two theorems will extend this result for the entire hybrid dynamics. %We will fix the values for $\epsilon, r$ and the gains $K_p, K_d$ obtained from the previous lemmas. Specific value of $\epsilon$, and also other constants like $L_q$ are picked in Appendix.

For simplicity of notation, we will denote the output coordinates as $\eta: = (e, \dot e)$, %, where $$(\eta,z) \in \Phi(\B_r(x^*) \cap \Guard).$$ 
and shift $z^*$ to zero. We will also drop the subscript $0$ from the initial states $\eta_0,z_0$\footnote{Since the ensuing Lemma and Theorems are established via \poincare maps, there is no confusion.}. 
Since we are analyzing periodic orbits, we will be using \poincare maps defined as %: $$\pcare: \Phi(\Guard) \to \Phi(\Guard)$$ obtained from the flow:
\begin{align}\label{eq:pcare}
\pcare(\eta, z) := \phi_{T(\eta,z)}(\ResetMap(\eta,z)),
\end{align}
where $T$ is the time-to-impact function \eqref{eq:dwelltime} (which is well defined based on \eqref{eq:timeineq}). $\pcare$ has two components $\pcare_\eta, \pcare_z$ corresponding to the coordinates $\eta,z$ respectively.
We can also establish some of the properties of impact maps.  We know that $\ResetMap$ can be separated into two components $(\ResetMap_\eta, \ResetMap_z)= \ResetMap$ corresponding to $\eta$ and $z$. By local Lipschitz continuity and hybrid invariance conditions, we have the following:
\begin{align}\label{eq:szeroineq}
%&  |\ResetMap(\eta, z) - \ResetMap(0,z)|   \leq  L_{\ResetMap} |\eta| \nonumber \\
%|	\ResetMap_\eta(\eta, z)  - \ResetMap_\eta( 0, z)| & \leq  L_{\ResetMap} |\eta| \nonumber \\
| z(0) - z_s(0) | = |\ResetMap_z(\eta, z) - \ResetMap_z(0,z)|   & \leq  L_{\ResetMap} |\eta|  \\
| z(0) - z^*(0) | = |\ResetMap_z(\eta, z) - \ResetMap_z(0,0)| & \leq  L_\ResetMap (|\eta| + | z |),\nonumber
\end{align}
where $L_{\ResetMap}$ is the Lipschitz constant. Since $\ResetMap_\eta(0,z)=0$, %We can obtain inequalities stricter than the one obtained in \eqref{eq:esetzineq} for $\eta(0)$:
\begin{align}\label{eq:eszeroineq}
%\left | \begin{bmatrix}
%e(0) \\ \dot e(0)
%\end{bmatrix} \right | % = & |\eta(0)| =   |\ResetMap_\eta(\eta, z)| \nonumber \\
|(e(0), \dot e(0))| =  |	\ResetMap_\eta(\eta, z)  - \ResetMap_\eta(0, z)| \leq  L_{\ResetMap} |\eta|.
\end{align}

There is an elegant relationship between the time-to-impact functions $T,T_\rho$ and the \poincare maps $\pcare, \rho$ that can be utilized to prove the main theorem. This is given in the form of the following lemma.
%\gap
\begin{lemma}\label{lm:timepcarelemma}{\it 
Let the system \eqref{eq:hsoutputzero} be given, and let the desired configuration $q^a_d$ be chosen such that Assumptions \ref{ass:phidiff}-\ref{ass:by} are satisfied. 
Given that $\Orbit_z$ is an LES periodic orbit of the HZD transverse to $\Phi(\ZD{} \cap \Guard)$,
% and if the desired trajectories $q^a_d$, phases $ \tau^s,\tau^t$ satisfy Assumptions \ref{ass:timebasedphase}-\ref{ass:phasematch}, 
then there exist large gains $k_p, k_d$, small enough $r>0$, and corresponding $C_T, C_P, d_T, d_P>0$ such that for all $(\eta,z)\in \B_r(0,z^*) \cap \Phi(\Guard)$, %$(e, \dot e, z) \in \B_r$
% such that for sufficiently large gains $K_p, K_d$, there exist constants $C_T, C_P, d_T, d_P>0$ that satisfy
		\begin{align}\label{eq:timetoimpactandpcaremap}
		\!\!|T(\pcare(\eta,z)) - T_\rho(z)| & \leq C_T |\eta|  + d_T  \\
		\label{eq:timetoimpactandpcaremap2}
		| \pcare_z(\eta,z) - \rho(z)|  & \leq C_P |\eta| + d_P.
		\end{align}
		In addition, the gains $k_p, k_d$ can be further increased and $r$ can be further decreased in such a way that $C_T, C_P$ remain constant and $d_T, d_P$ decrease (with increasing $k_p, k_d$). 
		%In other words, $d_T, d_P$ can be decreased to any arbitrary value by appropriate choice of the gains and the neighborhood.%$x_0 \in \B_r(x^*) \cap \Guard$.
	}
\end{lemma}
%\gap
Lemmas \ref{lm:boundedness}-\ref{lm:timepcarelemma} yield $r$ and the PD gains $k_p,k_d$, which yield the inequalities \eqref{eq:timetoimpactandpcaremap}, \eqref{eq:timetoimpactandpcaremap2}. %Having defined the \poincare maps $\pcare$, $\rho$, 
We will now state the following well known result for periodic orbits \cite{veer2017poincare}: 
%\gap
\begin{theorem}\label{thm:theoremfromveer}{\it
		Given the set of points $\Orbit$ obtained from the embedding $\iota$ of the periodic orbit $\Orbit_z$ \eqref{eq:periodicorbitreduced} of the hybrid zero dynamics $\mathcal{Z}$ \eqref{eq:hzds}, the following are equivalent:
		\begin{itemize}
			\item[1.] $\Orbit=\iota(\Orbit_z)$ of the hybrid system $\HybridSystem$ \eqref{eq:hsoutputzero} is  LEUB. %exponentially ultimately bounded.
			\item[2.] $(0,0)= \Phi(x^*)$ of the map $\pcare$ \eqref{eq:pcare} is LEUB. %exponentially ultimately bounded.
	\end{itemize}}
\end{theorem}
%\gap
Theorem \ref{thm:theoremfromveer} will be used to establish the main theorem of the paper that follows. We will not be proving this theorem, since it is directly obtained as a consequence of \cite[Theorem 1]{veer2017poincare} (by simply replacing the class $\classK$ function of the disturbance with a constant $d$). The proof of \cite[Theorem 1]{veer2017poincare} is for a stronger property--input-to-state stability (ISS)--and similar results can be derived for boundedness. \addblue{We can establish that the ultimate bound, given by $d$ in \eqref{eq:expobounded}, can be decreased to an arbitrarily small value by choosing appropriate $k_p, k_d$. As $d \to 0$, the resulting orbit of the closed loop system coincides with $\Orbit$. %Therefore the extension of Theorem \ref{thm:theoremfromveer} for PD controlled robotic systems is established by choosing a small enough neighborhood and large enough gains, thereby decreasing $d$.
Therefore, Theorem \ref{thm:theoremfromveer} will be used to establish boundedness of periodic orbits via \poincare maps.
}

\newsec{Main theorem.} We will now present the main theorem. % of the paper. 

%\gap
\begin{theorem}\label{thm:main}
	{\it 
	Let the system \eqref{eq:hsoutputzero} be given, and let the desired configuration $q^a_d$ be chosen such that Assumptions \ref{ass:phidiff}-\ref{ass:by} are satisfied. 
		If $\Orbit_z$ is an LES periodic orbit of the hybrid zero dynamics $\mathcal{Z}$ \eqref{eq:hzds} transverse to $\Phi(\ZD{} \cap \Guard)$, 
% 		is LES with the desired trajectories $\q^a_d$ and the phases $\tau^s, \tau^t$ satisfying Assumptions \ref{ass:timebasedphase}-\ref{ass:phasematch}, 
		then for sufficiently large enough gains $k_p, k_d>1$, $\Orbit$ is an LEUB periodic orbit of the full order hybrid dynamics $\HybridSystem$ \eqref{eq:hsoutputzero}.
	}
\end{theorem}
%\gap

We will prove Lemmas \ref{lm:boundedness}-\ref{lm:timepcarelemma}, and also Theorem \ref{thm:main} in the next section.

\section{Proofs of Lemmas \ref{lm:boundedness}-\ref{lm:timepcarelemma} and Theorem \ref{thm:main}}\label{sec:proofs}

\begin{pf}[of Lemma \ref{lm:boundedness}] \label{proof:lemmaboundedness}
		We consider the following Lyapunov candidate:
		\begin{eqnarray}
		\label{eq:Lyapunovcandidatereduceddimensions}
		V_e(e,\dot e,\q)  & = & V_0(e,\dot e,\q) +  V_{c} (e,\dot e,\q) \\
		\label{eq:Vo}
		V_0 (e,\dot e,\q)  & =& \frac{1}{2}\begin{bmatrix}
		e \\ \dot e 
		\end{bmatrix}^T  \begin{bmatrix}
		K_p & \0 \\
		\0 & \Ba^T A_e(\q) D_e(\q) \Ba
		\end{bmatrix} \begin{bmatrix}
		e \\ \dot e 
		\end{bmatrix} \\
		\label{eq:Vc}
		V_{c} (e,\dot e,\q)  & = &  \alpha(e) e^T  \Ba^T A_e(\q) D_e(\q) \Ba \dot e  \\
		\label{eq:alpha}
		\alpha(e)  & = &  \frac{k_0}{1 + |e|} = \frac{k_0}{1 + \sqrt{e^T e}}.
		\end{eqnarray}
		It can be verified that $V_e$ is positive definite. The addition of the cross terms $V_c$ does not affect the positive definiteness as long as $k_0$ is sufficiently small. For example, we can pick $k_0$ that satisfies
		\begin{align}\label{eq:kappazerofirstpickunderactuated}
		k_0 \leq \frac{\sqrt{\|K_p\|\|\Ba^TA_e D_e \Ba\|}}{\|\Ba^TA_e D_e \Ba\|}. % = \frac{(\|K_p\|\underline{c}_{ad})^{\frac{1}{2}}}{\bar {c}_{ad}}.
		\end{align}
		Therefore, we can choose 
		\begin{align}\label{eq:k0}
		k_0 = \frac{\sqrt{k_p}}{N}, \:\: {\rm{where}} \:\: N > \|\Ba^TA_e D_e \Ba\|^{\frac{1}{2}}. 
		\end{align}
		We will be picking a larger value for $N$ later on. Choosing this value of $k_0$ also helps in separating $k_p$ from the positive definite matrix in \eqref{eq:Vo} to obtain the following bounds on $V_e$:
		\begin{align}\label{eq:boundsforve}
		\lmin \left |  \begin{bmatrix} \sqrt{k_p} e \\ \dot e \end{bmatrix}   \right |^2 \leq V_e \leq  \lmax \left |  \begin{bmatrix} \sqrt{k_p} e \\ \dot e   \end{bmatrix}   \right |^2,
		\end{align}
		for some positive constants $\lmin$, $\lmax$ that do not depend on $k_p$. This type of inequality is useful for realizing desirable convergence rates for $V_e$ (see \cite[(10)-(22)]{TAC:amesCLF} for a similar formulation).
		%The constant $k_p$, which is the proportional gain, was separated from $\lmax, \lmin$ to indicate the dependency of upper bound of $V_e$ on $k_p$. %, which are dependent on the constants $k_0,k_p$. % = \max \{ 1, k_p, c_a \}$.
			
%		The goal is show that the growth of $V_t$ is finite in a finite interval $[0,T_{\max}]$ irrespective of the values of $k_p,k_d$. Therefore we will evaluate the derivatives of $V_t,V_0,V_c$ one by one below.
		\newsec{$V_e$ of the initial states.} 	It is important to determine the sublevel sets of $V_e$ that contains all the possible initial values. Since $\Orbit_z$ is locally stable in HZD, we can pick the same neighborhood radius $r$ (or maybe even smaller) for the initial states $(e_0, \dot e_0, z_0)$. With this initial state we know from \eqref{eq:szeroineq} and \eqref{eq:eszeroineq} that
				\begin{align}\label{eq:remark5second}
		|(e(0), \dot e(0))| \leq  &  L_\ResetMap | (e_0,\dot e_0)| \leq L_\ResetMap r.
		\end{align}
% 		where $L_\ResetMap$ is the local Lipschitz constant of the impact map (in transformed coordinates). 
		Therefore, given the initial state, the maximum possible value of $V_e$ is $  \lmax k_p  L_\ResetMap^2 r^2$. %, and the maximum possible value of the states $e_t, \dot e_t$ in this level set is given by 
%		\begin{align}
%			|e_t|_{\max} := 3 L_\ResetMap r, |\dot e_t|_{\max} := 
%		\end{align} 
%		These values will be important when choosing the 
		This will be useful for determining the gains of the controller later on. We will now solve for the dynamics of $V_e$.
		
		\newsec{Derivative of $V_0$.} By solving for the derivative of $V_0$, we get the following:
		\begin{align}
		\dot V_0 &=  e^T K_p \dot e  + \frac{1}{2} \dot e^T \Ba^T (\dot A_e D_e + A_e \dot D_e) \Ba \dot e  + \dot e^T \Ba^T A_e D_e \Ba \ddot e, \nonumber 
		\end{align}
		and after substituting \eqref{eq:eliminatederrordynamics} and \eqref{eq:pdtrackingunderactuated} and using Propositions \ref{proposition:dece}, \ref{prop:schurpde}, we have the final inequality as
		\begin{align}\label{eq:amigo1}
		\dot V_0 =&  \frac{1}{2} \dot e^T \Ba^T (\dot A_e D_e + A_e \dot D_e) \Ba \dot e  - \dot e^T \Ba^T A_e( C_e \begin{bmatrix} \dot z_1 \\ \dot e \end{bmatrix} + G_e ) \nonumber \\
		 & - k_p \dot e^T  (B_e - \1) e - k_d \dot e^T  B_e \dot e  \nonumber \\
		 \leq & \frac{c_u}{2} \left |\begin{bmatrix} \dot z_1 \\ \dot e\end{bmatrix} \right | |\dot e|^2 + c_u \left | \begin{bmatrix} \dot z_1 \\ \dot e\end{bmatrix}\right |^2 |\dot e| + c_u |\dot e| \nonumber \\
		 & - k_p \dot e^T  (B_e - \1) e - k_d \dot e^T  B_e \dot e .
		\end{align}
		
		\newsec{Derivative of $V_c$.} By solving for the derivative of $V_c$, we have
		\begin{align}\label{eq:amigo2}
		\dot V_c =& \dot \alpha e^T \Ba^T A_e D_e \Ba \dot e + \alpha \dot e^T \Ba^T A_e D_e \Ba \dot e  \nonumber \\
		& +  \alpha e^T \Ba^T (\dot A_e D_e + A_e \dot D_e) \Ba \dot e  + \alpha e^T \Ba^T A_e D_e \Ba \ddot e   \nonumber \\
		\leq & 2 \alpha c_u  |\dot e|^2  + \alpha c_{u} \left| \begin{bmatrix} \dot z_1 \\ \dot e \end{bmatrix} \right| | e| |\dot e| + \alpha c_u  \left| \begin{bmatrix} \dot z_1 \\ \dot e \end{bmatrix} \right|^2 |e|+\alpha c_u |e| \nonumber \\
		&  - k_p \alpha e^T B_e  e - k_d \alpha e^T B_e \dot e,
		\end{align}
		where we have used Propositions \ref{proposition:dece}, \ref{prop:schurpde}, and some of the properties of $\alpha(e)$:
		\begin{alignat*}{5}
		% 	\alpha(e) &= \frac{k_0}{1 + |e|}  \nonumber \\
		 |\alpha(e)|  & \leq   k_0 ,	\quad & 	 |\alpha(e)|   |e| & \leq  k_0 , &\nonumber \\
		 | \dot \alpha(e) |  & \leq  k_0 |\dot e| ,\quad  & 	| \dot \alpha(e)||e|  & \leq  k_0 |\dot e|. &
		\end{alignat*}
		In addition, we note from \eqref{eq:zerocoordinates}, Property \ref{prop:1} and Assumptions \ref{ass:phidiff}, \ref{ass:by} that 
  $B_e' \dot z_1  = D_{11}^{-1} ( z_2  - D_{12}  J^{-1} \dot e)$,
% |\dot z_1|  =  | D_{11}^{-1}(\q) ( z_2  - D_{12} (\q) \dq^a)| \leq  c_u |z_2| + c_u^2 ( |\dot e| + c_q). \nonumber
% \end{align}
which yields 
\begin{align}\label{eq:zdotineq}
\left |\begin{bmatrix} \dot z_1  \\ \dot e\end{bmatrix} \right|  \leq  c_q \left | \begin{bmatrix} z_2    \\ \dot e\end{bmatrix} \right |,
\end{align}
for some $c_q>0$. This can be substituted for $\dot z_1, \dot e$ in the above equations.
%		\newsec{Derivative of $V_z$.} Taking the derivative of $V_z$ yields
%		\begin{align}
%		%	\dot V_z = 2 {q^u}^T \dot {q^u} + 2 \dq^T D(\q) B_c B_c^T ( \dot D(\q) \dq +  D(\q) \ddq ),
%		\dot V_z %=  & 2 (z - z^*(t))^T ( \psi(e,\dot e,z) - \dot z^*(t) ) \nonumber \\
%		  = &  \frac{\partial V}{\partial t} + \frac{\partial V}{\partial z} \psi(e,\dot e,z) \nonumber \\
%		%		 = & 2 {z_1}^T \dot  q^u + 2 \dq^T D(\q) B_c B_c^T (\dot D \dq - C\dq - G) \nonumber \\
%%		\leq & 2 |\tilde z|  (|\dot z_1| + 2 c_u |\dq|^2  + c_u + k_q),
%%		= & 2 (z - z^*(t))^T ( \psi(0,0,z) - \dot z^*(t) ) \nonumber \\
%%		&  + 2 (z - z^*(t))^T ( \psi(e,\dot e,z) - \psi(0,0,z)), 
%		\leq & - c_z |z - z^*(t) |^2 +  \frac{\partial V}{\partial z} (\psi(e,\dot e,z) - \psi(0,0,z)),
%		\end{align}
%		%which can be substituted for \eqref{eq:eom-general} to yield
%		%\begin{align}
%		%	\dot V_z = 2 {q^u}^T \dot {q^u} + 2 \dq^T D(\q) B_c B_c^T ( \dot D(\q) \dq  - C_l \dq  - G_1 )	,
%		%\end{align}
%		which does not contain the input $u$ (due to the passivity of the dynamics of $z$). Note that $\psi$ (obtained from \eqref{eq:psidynamics}) depends on the state based outputs $e, \dot 
%		We can obtain the total derivative $\dot V_t$ by adding $\dot V_e + \dot V_z$. Considering $\dot V_e$ first, we have
\addblue{
We obtain the total derivative as
\begin{align}\label{eq:totalvedot}
\dot V_e  \leq & - \frac{\alpha}{2} \begin{bmatrix}
e \\ \dot e
\end{bmatrix}^T \underbrace{\begin{bmatrix}
k_p (B_e + B_e^T)  &  \frac{ k_p(B_e^T - \1)  +  \alpha k_d B_e }{ \alpha}  \\
\frac{ k_p (B_e - \1)  +  \alpha k_d B_e^T }{ \alpha} &  \frac{k_d (B_e + B_e^T) }{\alpha} 
\end{bmatrix}}_{\Lambda} \begin{bmatrix}
e \\ \dot e
\end{bmatrix} \nonumber \\
& + \alpha c_u |z_2| |e| |\dot e| + c_u (\alpha |e| + |\dot e|)  (|z_2|^2 + 1) \nonumber \\
& + 2 c_u ( \alpha + \alpha |e| + |\dot e| + |z_2|) |\dot e|^2,
% & + c_u \left ( \left | \begin{bmatrix} \dot z_1 \\ \dot e \end{bmatrix}\right | + 1 + 2\alpha \right )|\dot e|^2. 
\end{align}
where $c_u$ is redefined to absorb $c_q$. The control gains $k_p, k_d$ must be picked in such a way that $V_e$ is decreasing. For $|z_2|$, we will pick the maximum of $z$ in a compact neighborhood (tube) of $\Orbit_z$. Assume that this maximum is $b>0$ (say). Similarly, the maximum possible values for $|\dot e|, |e|$ in the sublevel set of $V_e$ are given by
\begin{align}\label{eq:emaxedotmax}
|\dot e | \leq   \sqrt{k_p \frac{\lmax}{\lmin}}  L_\ResetMap  r =: |\dot e |_{\max} , \:  |e| \leq    \sqrt{\frac{\lmax}{\lmin}}  L_\ResetMap  r =: |e|_{\max} ,
\end{align}
which are obtained from the maximum value of $V_e$ for all possible initial states. Since $|\dot e |_{\max}$ depends on the proportional gain $k_p$, the derivative gain $k_d$ must be at least as high as $\sqrt{k_p}$. Therefore, choose $k_p =  \epsilon^2, k_d = \epsilon k, k_0 = \epsilon/k$ for some $\epsilon, k>1$. We require that $\Lambda\succ 0$ i.e.,
\begin{align}
 \begin{bmatrix}
     e \\ \dot e
    \end{bmatrix}^T \Lambda \begin{bmatrix}
     e \\ \dot e
    \end{bmatrix}  =  \begin{bmatrix}
     \epsilon e \\ k \dot e
    \end{bmatrix}^T \Lambda_{B_e} \begin{bmatrix}
     \epsilon e \\ k \dot e
    \end{bmatrix} > 0, 
\end{align}
for all $(\q,\dq) \in \cup_{\q\in \mathbb{N}_r} T_\q \Q$. This is true by Assumption \ref{ass:by}. Therefore, we have the following:
\begin{align}\label{eq:totalder}
\dot V_e  \leq & - \frac{\alpha}{2} \lambda_{B_e}  \left | \begin{bmatrix}
     \epsilon e \\ k \dot e
    \end{bmatrix} \right |^2
        % & + \alpha k_4 |\dot y|^2 + \alpha k_5 |y| |\dot y| 
         + \alpha k_1 |e| |\dot e| + \alpha k_2 |\dot e|^2 \\
         &  + k_3 (\alpha |e| + | \dot e|), \nonumber
		\end{align}
where $\lambda_{B_e}: = \lambda_{\rm{min}}(\Lambda_{B_e})$ is the minimum eigenvalue in the compact set $\mathbb{N}_r$, and 
\begin{align}
\label{eq:k1k2k3}
% & \Lambda = \begin{bmatrix}
%                                        k_p (B_e + B_e^T)  &  \frac{k_p(B_e - \1)  +  \alpha k_d B_e  - \alpha k_5 \1}{\alpha}  \\
%                                         \frac{k_p (B_e - \1)  +  \alpha k_d B_e  - \alpha k_5 \1}{\alpha} &  \frac{k_d (B_e + B_e^T)  - 2c_u c_z |\dot e|\1 - k_4 \1}{\alpha}
%                                     \end{bmatrix} \nonumber \\
&  k_1(|z_2|) =   c_u |z_2|   \nonumber \\
&  k_2(|e|, |\dot e|, |z_2|) = 2  c_u \left ( 1 + |e| + \frac{(|\dot e| + |z_2|)}{\epsilon k^{-1}(1 + |e|)^{-1}} \right )  \\
&  k_3(|z_2|) = c_u ( |z_2|^2 +  1  ). \nonumber
\end{align}
One half of the first summand in \eqref{eq:totalder} can be used to cancel the next two summands in the following manner:
\begin{align}\label{eq:ineqthreesumamnds}
 -  \frac{ \lambda_{B_e}}{4} |(\epsilon e, k \dot e)|^2  &+ k_1 |e| |\dot e| + k_2 |\dot e|^2 =  \\
 & - \begin{bmatrix}
                                                              |e| \\ |\dot  e|
                                                             \end{bmatrix}^T
  \begin{bmatrix} \frac{\epsilon^2 \lambda_{B_e}}{4}  & - k_1 \\
  - k_1 &    \frac{k^2 \lambda_{B_e}}{4}  - k_2    \end{bmatrix} \begin{bmatrix} |e| \\ |\dot  e| \end{bmatrix}, \nonumber
\end{align}
where the summands are collected together in the form of a matrix. We can replace $e, \dot e, z_2$ in $k_1, k_2$  with their maximum possible values, and by choosing a large enough $k$, we can ensure that \eqref{eq:ineqthreesumamnds} is negative. 
}

Note that \eqref{eq:ineqthreesumamnds} is satisfied even if $\epsilon$ is increased ($|\dot e| \leq |\dot e|_{\rm{max}}$ is canceled by $\epsilon$ in the denominator in \eqref{eq:k1k2k3}), and if the $k$ chosen does not satisfy \eqref{eq:kappazerofirstpickunderactuated} i.e., if $k \not > N$, then we simply increase it further. We will be using $\epsilon$ as a tunable gain for obtaining desirable convergence rates. We have that
% \begin{align}
%     \lmin \left |\begin{bmatrix} \epsilon e \\ \dot e \end{bmatrix} \right |^2 \leq \begin{bmatrix} e \\ \dot e    \end{bmatrix}^T     \Lambda \begin{bmatrix} e \\ \dot e    \end{bmatrix} \leq \lmax \left |\begin{bmatrix} \epsilon e \\ \dot e \end{bmatrix} \right |^2
% \end{align}
% where the same $\lmin, \lmax$ notations are used for convenience (by redefining). Given that $\Lambda \succ 0$, we have 
 \begin{align}\label{eq:veineqlambdad}
         \dot V_e & \leq  -  \frac{\alpha \lambda_{B_e} }{4} |(\epsilon e, k \dot e) |^2 + k_3 (\alpha |e| + | \dot e|) \nonumber \\
                  & \leq -  \frac{\alpha \lambda_{B_e} }{8} |(\epsilon e, \dot e) |^2 +   \frac{2 \alpha k^2_3}{ \lambda_{B_e} \epsilon^2} + \frac{2 k^2_3 k^2}{ \lambda_{B_e} \alpha} \nonumber \\
                  & \leq -2 \epsilon \lambda V_e + \frac{2 \alpha k^2_3}{ \lambda_{B_e} \epsilon^2} + \frac{2 k^2_3 k^2}{ \lambda_{B_e} \alpha},
        %  \frac{k_6^2}{2k_d}+\frac{\alpha k_6^2}{2k_p},
 \end{align}
where $\lambda := \lambda_{B_e}/(16 k (1+|e|_{\max}) \lmax)$. The above inequality is satisfied as long as $z$ remains in the tube. Let $T_\delta$ be the time when $z$ crosses this limit. Therefore, by using comparison lemma \cite[Lemma 3.4]{khalil2002nonlinear} in \eqref{eq:veineqlambdad}, we have that 
\begin{align}
    V_e(t) \leq e^{-2 \epsilon \lambda t} V_e(0) + \frac{k_4}{\epsilon^2}, \quad t\in [0,T_\delta],
\end{align}
where $k_4$ is obtained by collecting all the additional terms that are independent of $\epsilon$. We can express the above inequality in terms of the outputs as
\begin{align}\label{eq:outputsleub}    \left | \begin{bmatrix} e(t) \\ \dot e(t) \end{bmatrix} \right |^2 & \leq \epsilon^2 e^{- 2 \epsilon \lambda t} \frac{\lmax}{\lmin} \left | \begin{bmatrix} e(0) \\ \dot e(0) \end{bmatrix} \right |^2 + \frac{k_4}{\epsilon^2  \lmin} \nonumber \\
    \left | \begin{bmatrix} e(t) \\ \dot e(t) \end{bmatrix} \right | & \leq \epsilon e^{- \epsilon \lambda t} \sqrt{\frac{\lmax}{\lmin}} \left | \begin{bmatrix} e(0) \\ \dot e(0) \end{bmatrix} \right | + \sqrt{\frac{k_4}{\epsilon^2  \lmin}},
\end{align}
which completes the proof. 
\end{pf}
\begin{remark}{\it
		%For a positive $t>0$, increasing $\epsilon$ results in decreasing of the outputs at time $t$: $e_t(t), \dot e_t(t)$, but introduces a peak within the interval $(0,t)$. This peak is due to the multiplication of $\epsilon$ to the first term in \eqref{eq:finaltheorem1resulteiss}. Therefore, increasing of $\epsilon$ (gains) increases the peak. 
		We showed that increasing $\epsilon$ increases the convergence rate of the outputs. It is important to note that this also results in changing shape of the Lyapunov level sets of $V_e$. The peak velocity $|\dot e|_{\rm{max}}$ increases with increasing $\epsilon$. On the other hand, increasing $\epsilon$ does not increase $|e|_{\rm{max}}$. Therefore, Assumptions \ref{ass:phidiff} and \ref{ass:by} are still valid. We will now %show that, by appropriately choosing $r$ and $\epsilon$ the above result can be extended for $[0,T]$. This will be the main body of the proof of 
		prove Lemma \ref{lm:zboundedness}. %, which is given below.
	}
\end{remark}
% \gap
\begin{pf}[of Lemma \ref{lm:zboundedness}]
\addblue{
Firstly, we note that by the differentiability of $\Phi$ (diffeomorphism):
\begin{align}\label{eq:qedifference}
    |\q(t) - \q^*(t)| \leq L_q ( |z_1(t) - z_1^*(t)| + |e(t)|),
\end{align}
where $L_q$ is the local Lipschitz constant (for the tube of radius $r$ around the orbit). 
%Secondly, from the definition of error $e$, we have that
%\begin{align}\label{eq:qederivative}
  %     \dq(t) - \dq^{*}(t) = & J_e^{-1}(\q(t)) \begin{bmatrix} \dot z_1(t) - \dot z_1^*(t) \\ \dot y(t) \end{bmatrix}\\
  % & + (J_e^{-1}(\q(t))-J_e^{-1}(\q^*(t)))\begin{bmatrix}    \dot z_1^*(t) \\ \0 \end{bmatrix}. \nonumber 
%\end{align}
By using the definition of the zero coordinates, we have that
\begin{align}\label{eq:zdifference}
    \dot z_1(t) - \dot z_1^*(t) = & B_e'(\q(t))^{-1} D_{11}(\q(t))^{-1}  \lbrack z_2(t) \nonumber \\
			    & \left .- D_{12}(\q(t)) J(\q(t))^{-1} \dot e(t) \right \rbrack \\
     & -  B_e'(\q^*(t))^{-1} D_{11}(\q^*(t))^{-1} z^*_2(t). \nonumber
\end{align}
Here $B_e'$ is given by \eqref{eq:bebeprime}, and $\q^*(t), z_1^*(t)$ correspond to the trajectory on $\Orbit$.}
% From \eqref{eq:qederivative}, we have that
% \begin{align}
%     \begin{bmatrix}
%     \dot z_1(t) - \dot z_1^*(t) \\
%     \dq^a(t) - \dq^{a*}(t)
%     \end{bmatrix} = &J_e^{-1}(\q(t)) \begin{bmatrix} \dot z_1(t) - \dot z_1^*(t) \\ \dot y(t) \end{bmatrix}\\
%   & +(J_e^{-1}(\q(t))-J_e^{-1}(\q^*(t)))\begin{bmatrix}    \dot z_1^*(t) \\ \0 \end{bmatrix}, \nonumber
% \end{align}
% which yields 
% \begin{align}
%      \dq^a(t) - \dq^{a*}(t) =  B^T \dq(t) -  B^T \dq^{*}(t),
% \end{align}
% \begin{align}
%       \dq^a(t) - \dq^{a*}(t) = & B^T J_e^{-1}(\q(t)) \begin{bmatrix} \dot z_1(t) - \dot z_1^*(t) \\ \dot y(t) \end{bmatrix}\\
%   & +B^T (J_e^{-1}(\q(t))-J_e^{-1}(\q^*(t)))\begin{bmatrix}    \dot z_1^*(t) \\ \0 \end{bmatrix}, \nonumber 
% \end{align}
% which can be used to substitute for $\dq^a(t)$ in \eqref{eq:zdifference}. The resulting expression will have $\dot z_1(t) - \dot z_1^*(t)$ on both the sides. 
\addblue{By collecting the common terms, we obtain the following inequality:
\begin{align}\label{eq:z1dotdifference}
    |\dot z_1(t) - \dot z_1^*(t)| \leq  & c_z ( |z_2(t) - z^*_2(t)| + |\q(t) - \q^*(t)| + |\dot e(t)|), 
\end{align}
for some $c_z>0$. We can have a similar inequality for the derivative of $z_2(t) - z_2^*(t)$:
\begin{align}\label{eq:z2dotdifference}
   |\dot z_2(t) - \dot z_2^*(t)| \leq  c_z( &  |\dot z_1(t) - \dot z_1^*(t) | + |\dot z_1(t) - \dot z_1^*(t) |^2  \\
    & + |\dot e(t)| + |\dot e(t)|^2+ |\q(t) - \q^*(t)| ), \nonumber
\end{align}
for, perhaps, a larger value for $c_z>0$. By using \eqref{eq:qedifference}, \eqref{eq:z1dotdifference} and \eqref{eq:z2dotdifference}, we establish that
% \eqref{eq:zdifferentdotineq}.
\begin{align}\label{eq:zdifferentdotineq}
%|\dot z_1(t) - \dot z_1^*(t)| & \leq c_z ( |z(t) - z^*(t)| + |\eta(t)|) \nonumber \\
|\dot z(t) - \dot z^*(t)| & \leq c_z \left ( \left | \begin{bmatrix} z(t) - z^*(t) \\ \eta(t)  \end{bmatrix} \right | +  \left | \begin{bmatrix} z(t) - z^*(t) \\ \eta(t)  \end{bmatrix} \right |^2 \right ),
% |\dot z(t) - \dot z^*(t)| & \leq c_z ( |z(t) - z^*(t)| + |\eta(t)|),
\end{align}
for some $c_z>0$ (possibly larger than the previously determined $c_z$). We have used $\eta(t) := (e(t), \dot e(t))$ for convenience.
% Similarly, we have that
% \begin{align}
%     %z_2(t) - z_2^*(t)  \leq & D_z(\q(t)) J_e^{-1}(t,\q(t))  \begin{bmatrix}\dot z_1(t) \\ \dot y(t)  + \frac{\partial y_d(t,\q(t))}{\partial t} \end{bmatrix} \nonumber \\
%   % & - D_z(\q^*(t)) J_e^{-1}(t,\q^*(t))  \begin{bmatrix}\dot z^*_1(t) \\   \frac{\partial y_d(t,\q^*(t))}{\partial t} \end{bmatrix}.
%   z_2(t) - z_2^*(t)  \leq & D_z(\q(t)) \dq(t) - D_z(\q^*(t)) \dq^*(t) .
% \end{align}
% To establish the second inequality of Lemma \ref{lm:zlemma}, we first verify that $z_2$ remains bounded by $b$ in the interval $[0,T_\delta]$.  %The key idea is to choose an appropriate $r$ for a fixed value of $\epsilon$. 
We substitute \eqref{eq:zdifferentdotineq} in the following inequality:
\begin{align}
| z (t) - z^*(t) | \leq & | z(0) - z^*(0) |  + \int_0^{t} \left | \dot z(s) - \dot z^*(s)    \right |  ds \nonumber \\
    \leq & | z(0) - z^*(0) | +  c_z  \int_0^{t} (|\eta(s)| + |\eta(s)|^2) ds \nonumber \\
     & + c_z \int_0^{t} (|z(s) - z^*(s)| + |z(s) - z^*(s)|^2 ) ds. \nonumber
\end{align}
For a tube of radius $r$ around $\Orbit_z$, we have that $|z(t)| , |z^*(t)|\leq b$. Therefore, by substituting $b$ and the expression for the outputs $\eta(t)$, we have that
\begin{align}
| z (t) - z^*(t) | \! \leq & |z(0) - z^*(0) |  + \frac{c_z\epsilon \lmax}{2\lambda \lmin} |\eta(0)|^2 + \frac{c_z k_4 T_\delta }{\epsilon^2 \lmin}   \nonumber \\
                                        & + \frac{c_z}{\lambda} \sqrt{\frac{\lmax}{\lmin}} |\eta(0)| + c_z\sqrt{\frac{k_4}{\epsilon^2 \lmin}} T_\delta  \nonumber \\
                                        & + c_z(1\!+\!2b)\! \int_0^{t} \! |  z(s) -  z^*(s)     |  dt.
\end{align}
By using Gronwall-Bellman inequality \cite[Lemma A.1]{khalil2002nonlinear}, we obtain
\begin{align}\label{eq:zdifferenceaftergronwallbellman}
\! | z (t) - z^*(t) |\! \! \leq &  c_z \left ( \! |z_0 | \!  + \! \epsilon |\eta_0|^2 + |\eta_0| + \frac{T_\delta }{\epsilon^2 }    + \frac{T_\delta}{\epsilon}   \right ) e^{c_z(1+2b)t},
\end{align}
where the pre-impact state $(\eta_0,z_0)$ is substituted and $c_z$ is appropriately redefined. We establish the first part of Lemma \ref{lm:zboundedness} by substituting $z_0=z^*=0$, $\eta_0=0$ and appropriately choosing $C_k, C_t$. } 

To establish the second part of Lemma \ref{lm:zboundedness}, we first note that the constant terms in \eqref{eq:zdifferenceaftergronwallbellman} decrease with increasing $\epsilon$. %(also due to decreasing $r$, see Remark \ref{rm:repsilon}):
Therefore, $z(t)$ gets closer to $\Orbit$ with decreasing $r$ and increasing $\epsilon$, implying that the states are bounded for $[0,T_\delta]$. Since $T_\delta$ is the time when $|z_2(t)|$ crosses its bound $b$, i.e., leaves the compact neighborhood around $\Orbit_z$, we can use the steps followed in \cite[Theorem 3.5]{khalil2002nonlinear}. $T_\delta$ is stretched to an arbitrary constant $T_\delta = T_{\max} > T^*$ (say), %$T(0,0,z^*)$ with a sufficiently large $\epsilon$ (say $\epsilon_1$).
%Assume that the value of $\epsilon$ required is $\epsilon_1 >0$, which is greater than the $\epsilon$ chosen in proof of Lemma \ref{lm:boundedness}. 
and by choosing $\epsilon$ (say $\epsilon_1$), and $r$ (say $r_1$ that is smaller than previously chosen), we ensure that Lemma \ref{lm:boundedness} is valid for the entire $[0,T_{\max}]$. 
%Therefore \eqref{eq:zerrordynamics} is valid for any arbitrary interval $[0,T_\delta]$. 
With this result, the next steps are similar to \cite[Proof of Lemma 2]{veer2017poincare}, where we use implicit function theorem (in function spaces) to establish bounds for the time-to-impact function \eqref{eq:dwelltime}. We know that on the nominal orbit, we obtain the period $T^*$ as 
\begin{align}\label{eq:timetoimpactnominal}
T^* = \inf \{ t>0	 : h (\Phi^{-1}(0,0,z^*(t))) = 0\},
\end{align}
which is well defined. By transversality condition (see \cite[A.7]{veer2017poincare}), any small perturbation of the trajectories from the nominal orbit in the guard function $h (\Phi^{-1}(\mu_1(t),0,z^*(t)+\mu_2(t)))$ will still yield a well defined time-to-impact function, as long as the perturbations $\mu_1(t), \mu_2(t)$ are small\footnote{$h$ is not dependent on velocity $\dot e$, and hence the second argument being zero still yields the time-to-impact function.}. Pick $\mu_1(t) = e(t)$, $\mu_2(t) = z (t)- z^*(t)$, and the resulting guard function yields \eqref{eq:dwelltime}. 
We can quantify $\mu_i$'s by using function norms. Denote the functional norms as 
%\begin{align}
%\|(e,\dot e)\|_1 &:= \int_{0}^{T_{\max}}  |(e(t),\dot e(t))| dt \nonumber \\
%\|z - z^*\|_1 &:= \int_{0}^{T_{\max}} | z(t) - z^*(t)| dt,
%\end{align}
%which can be evaluating by using \eqref{eq:finalzineq} and \eqref{eq:finaleinequality}. 
%where $T_m$ is a constant (see proof of Lemma \ref{lm:zboundedness}) chosen to be much larger than $T^*$.
\begin{align}
\|e\|_\infty &:= \sup_{t \in [0,T_{\max}]} |e(t)| \nonumber \\
%\|\dot e\|_\infty &:= \sup_{t \in [0,T_{\max}]} |\dot e(t)| \nonumber \\
\|z - z^*\|_\infty &:= \sup_{t \in [0,T_{\max}]} | z(t) - z^*(t)|.
\end{align}
Both $\|e\|_\infty$ and $\|z - z^*\|_\infty$ can be evaluated from \eqref{eq:outputsleub} and \eqref{eq:zdifferenceaftergronwallbellman} respectively.
% respectively for the time interval $[0,T_{m}]$. 
%By implicit function theorem, 
Since the initial state is $(0,0,z^*)$, we know that by increasing $\epsilon$,
%$\|(e,\dot e)\|_1$, and $\|z-z^*\|_1$ 
$\|e\|_\infty$ %, %$\|\dot e\|_\infty$, 
and $\|z-z^*\|_\infty$ 
can be decreased.
% , $T(0,0,z^*)$ is well defined and finite. %, $T_B$ is well defined. Moreover $T  = T_B$ when 
%By implicit function theorem, we know that by reducing $\|(e,\dot e)\|_1$ and $\|z-z^*\|_1$ to a small enough value, $T$ is well defined.
%We complete the proof by %using Remark \ref{rm:epsiloninfty} %small enough $r$, 
Given $\delta >0$, if $\delta > T_{\max} - T^*$, we will first pick a smaller $\delta_1 >0$ that ensures that $T^* + \delta_1 < T_{\max}$.
Therefore, given $\delta_1 >0$, we can obtain a larger $\epsilon>\epsilon_1$ (say $\epsilon_2$) such that \eqref{eq:tstartdelta} is satisfied.

We will now establish the third part of Lemma \ref{lm:zboundedness}. Similar to $T_{\max}$, we can manually set some $T_{\min} > 0$ such that $T_{\min} < T^* < T_{\max}$\footnote{In \cite[Lemma 1]{TAC:amesCLF}, $T_{\min}, T_{\max}$ were chosen to be $0.9 T^*, 1.1 T^*$ respectively.}. %Also choose $T_\delta = T_{\max}$ in \eqref{eq:zerrordynamics}, and then obtain $\epsilon>1$ (say $\epsilon_1$) 
% Also choose an $\epsilon$ (say $\epsilon_1$) in such a way that for zero initial state, $(0,0,z^*)$, $|z_2(t)| \leq b/2$ for all $t\in[0,T_{\max}]$\footnote{We will choose $b/2$ instead of $b$ to be conservative.}. % In other words, we are ensuring that outputs are exponentially convergent for the entire interval $[0,T_{\max}]$ . 
We choose $ \delta < \min\{T_{\max} - T^*,T^* - T_{\min}\}$, and the corresponding time-to-impact satisfies 
%\begin{align}\label{eq:tobound}
$T_{\min} < T(0,0,z^*) < T_{\max}$, %for an $\epsilon_1. 
%\end{align}
for $\epsilon \geq \epsilon_2$. Note that $\epsilon$ can be increased even further to decrease the bound on $e,z$. Accordingly, we can choose a small enough neighborhood (smaller than $r_1$) around $(0,0,z^*)$ in such a way that \eqref{eq:timeineq} is satisfied, while still satisfying the conditions of implicit function theorem and boundedness of $z(t)$. %$|z_2(T_{\max})| \leq b$ and
%\begin{align}\label{eq:timebounde}
%T_{\min} \leq T(e_0, \dot e_0,z_0) \leq T_{\max}.
%\end{align}
%Since we have fixed $T_\delta = T_{\max}$ we would also like to ensure that \eqref{eq:finalzineq}, \eqref{eq:finaleinequality} are still true for a nonzero $r$ i.e., for all $(e, \dot e, z) \in B_*(r, \epsilon_1)$ and for all $t \in [0,T_{\max}]$. Since increasing $\epsilon$ and %decreasing $r$ 
%decrease both \eqref{eq:finalzineq}, \eqref{eq:finaleinequality}, we can pick a sufficiently larger $\epsilon = \epsilon_2 > \epsilon_1$ %%and smaller $r$
%that still ensures \eqref{eq:tobound}. 
This completes the proof.
\end{pf}
% \gap
\begin{remark}\label{rm:epsiloninfty}{\it
		For the initial state $(0,0,z^*)$, we have that $\epsilon\to \infty$ $\implies$ the resulting evolution of $z(t)$ is identically equal to $z^*(t)$. %Even the outputs $e(t), \dot e(t)$ are identically zero for all $t\in [0,T^*]$. %	Henceforth, we will be denoting $C_1(\epsilon)$ to indicate its dependency on $\epsilon$.
		For the non-zero initial state $(e_0, \dot e_0, z_0)$, it can be verified that the norm $\|z - z^*\|_\infty$ does not increase with increasing $\epsilon$ if the neighborhood radius $r$ is chosen appropriately. For example, we can choose $r$ such that it is inversely proportional to $\sqrt{\epsilon}$ (see \eqref{eq:zdifferenceaftergronwallbellman}).
		In addition, the peak value of $|e(t)|$ only depends on $r$ (see \eqref{eq:emaxedotmax}).
% 		In addition, it can be verified that the norm $\|e\|_\infty$  does not increase with increasing $\epsilon$ even for non-zero initial state $(e_0, \dot e_0, z_0)$. 
		This will be useful for increasing $\epsilon$ to arbitrarily large values later on. 
		We will now prove Lemma \ref{lm:timepcarelemma}.
	}
\end{remark}

% \gap
\begin{pf}[of Lemma \ref{lm:timepcarelemma}]
We will first study the progression of $\eta(t)$ till the end of the step $t=T$. By replacing $T$ with its extreme values $T_{\min},T_{\max}$, we have the following from \eqref{eq:outputsleub}:
	\begin{align}\label{eq:etainequality}
	|\eta(T)| \leq &  \beta(\epsilon)  |\eta| +  d_\eta(\epsilon), 
	\end{align}
	where the subscript $0$ was omitted for convenience, and
	\begin{align}
	\beta(\epsilon) =  \sqrt{\frac{\lmax}{\lmin}} \epsilon e^{- \epsilon \lambda T_{\min}}, \quad 	d_\eta(\epsilon)  = \sqrt{\frac{k_4}{\epsilon^2 \lmin}}.
	\end{align}
Here the dependency of $\beta,d_\eta$ on $\epsilon$ is explicitly shown. Note that $\beta$ has an upper bound (denote it by $\bar \beta$) that is independent of $\epsilon$. 

We can obtain similar inequalities for the zero coordinates. Firstly, we know that $z(t)$ is the trajectory of the zero coordinate on the full order hybrid dynamics \eqref{eq:hsoutputzero} with the initial condition $(\eta,z)$ on the guard, and $z_s(t)$ is the trajectory on the HZD with the initial condition $z$ on the guard. %For a fixed $\epsilon$, we choose an $r$ such that the resulting difference $z(t) - z_s(t)$ yields:
%The next steps are similar to the steps followed to obtain the inequality \eqref{eq:zdifferenceaftergronwallbellman}.
For a fixed $\epsilon \geq \epsilon_2$, we know that we can pick an $r$ (see Remark \ref{rm:epsiloninfty}) in such a way that
\begin{align}\label{eq:zinequality}
|z(T)  - z_s(T)|	\leq C_z | \eta |   + d_z(\epsilon),
\end{align}
where $C_z, d_z$ are both obtained by collecting all the terms together. This is similar to the inequality \eqref{eq:zdifferenceaftergronwallbellman}. The rest of the steps are very similar to \cite[Proof of Lemma 1]{TAC:amesCLF}. We define an auxiliary time-to-impact function:
\begin{align}\label{eq:auxdwelltime}
T_B(\mu_\eta, \mu_z ,z) =\inf_{t > 0} \{ t : h(\Phi^{-1}(\mu_\eta, z_s(t) + \mu_z)) =0 \},
\end{align}
where $\mu_\eta \in \R^{2m}$, $\mu_z \in \R^{2l}$, and satisfies
\begin{align}\label{eq:TB}
|T_B(\mu_\eta, \mu_z ,z) - T_\rho(z)| \leq L_B (|\mu_\eta | + |\mu_z|),
\end{align}
where $L_B$ is the Lipschitz constant. In addition, $\mu_\eta, \mu_z$ are further defined to be
\begin{align}
\mu_\eta = \eta(t) |_{t = T(\eta,z)}, \quad \mu_z = z(t) - z_s(t) |_{t = T(\eta,z)},
\end{align}
which yields $T_B(\mu_\eta, \mu_z, z) = T(\eta,z)$. With this result, and by substituting \eqref{eq:etainequality}, \eqref{eq:zinequality},
and also the bound on $\beta(\epsilon)$
in \eqref{eq:TB}, %which is $\bar \beta$, 
we get the first result \eqref{eq:timetoimpactandpcaremap}. %We have the following conservative estimates of $C_T, d_T$:

The second result \eqref{eq:timetoimpactandpcaremap2} will follow exactly the steps in \cite[Proof of Lemma 1]{TAC:amesCLF} with the substitution of \eqref{eq:timetoimpactandpcaremap}. Define
\begin{align}\label{eq:C3}
C_{\psi} = \max_{T_{\min} \leq t \leq T_{\max}} | \psi (0,z_s(t)) |.
\end{align}
We therefore have
\begin{align}
|\pcare_z (\eta, z) - \rho(z) | \!\! \leq  & | z(0) - z_s(0)  + \nonumber \\
 &\left. \int_{0}^{T(\eta, z)} \hspace{-5mm} \psi(\eta(t), z(t)) - \psi (0,z_s(t)) dt \right | \nonumber \\
 &  +	\left | \int_{T(\eta, z)}^{T_\rho(z)} \hspace{-4mm} | \psi (0,z_s(t)) | dt \right |.
\end{align}
The first two terms on the RHS are substituted with \eqref{eq:zinequality}, and the last term can be replaced with \eqref{eq:C3} and \eqref{eq:timetoimpactandpcaremap}. This completes the proof.
\end{pf}
\noindent We are now ready to prove the main theorem. 
% For ease of notations we will make the following substitutions for some of the constants:
% \begin{align}\label{eq:creplacement}
% 	C : = \max \{\bar \beta + C_1, c_4, C_P, L_\rho + c_5 \gamma \},
% \end{align}
% which will simplify the ensuing derivations.
\begin{pf}[of Theorem \ref{thm:main}]
% We will be using Theorem \ref{thm:theoremfromveer}.2 to establish the main result of the paper. 
Since we know that exponential stability of $\Orbit_z$ yields a discrete time Lyapunov function $V_z(z)$ \eqref{eq:discreteVrho}, we need to construct a suitable Lyapunov function for the full order system.

\newsec{Lyapunov candidate for $\pcare$:}
Consider the following Lyapunov candidate for the \poincare map:
\begin{align}\label{eq:V}
V(\eta,z) = \sigma |\eta|^2  +  V_z(z),
\end{align}
where $\sigma>0$ is a tunable constant, and $(\eta,z) $ is the initial state (by a slight abuse of notations).

We will analyze the two terms in RHS of \eqref{eq:V} separately. We note in \eqref{eq:etainequality} that $\beta(\epsilon)$ can be reduced by increasing $\epsilon$. Hence, we can obtain $\bar \epsilon > \epsilon_2$ in such a way that $\beta(\epsilon) < 1/2$ (say) for all $\epsilon \geq \bar \epsilon$. We may have to increase $\epsilon$ further to decrease the ultimate bound later on. We have
\begin{align}\label{eq:etadifferencefinal}
 |\pcare_\eta(\eta,z)|^2 -   |\eta|^2 \leq  & - (1 -  \beta(\epsilon)^2) |\eta|^2  + d_\eta(\epsilon)^2.
\end{align}
\noindent By using \eqref{eq:discreteVrho}, we have the following:
\begin{align}\label{eq:zdifferencefinal}
V_z (\pcare_z(\eta,z)) &  -  V_z(z) \nonumber\\
 = &  V_z (\pcare_z(\eta,z)) - V_z(\rho(z))   + V_z(\rho(z)) - V_z(z) \nonumber \\
   \leq & c_4 |\pcare_z(\eta,z) -\rho(z)|.(|\pcare_z(\eta,z)| + |\rho(z)|)   - c_3 |z|^2 \nonumber \\
    \leq &  -   c_3   |z|^2 + c_4 ( C_P |\eta| + d_P(\epsilon))^2  \nonumber \\
    & + c_4 (L_\rho  + c_5 \gamma) (C_P  |\eta| + d_P(\epsilon)) |z|,
\end{align}
where $L_\rho$ is the Lipschitz constant of $\rho$. The remaining steps are similar to \cite[Eqns. (92)-(94)]{kolathaya2016parameter}, which yields
\begin{align}
V(\pcare(\eta,z)) - V(\eta,z) \leq - \begin{bmatrix}
|\eta| \\ |z|
\end{bmatrix}^T \Lambda_{\HybridSystem} \begin{bmatrix}
|\eta| \\ |z|
\end{bmatrix}  +  d(\sigma, \epsilon),
\end{align}
where
\begin{align}
& \Lambda_{\HybridSystem} =  \begin{bmatrix}
\sigma \frac{1 - 2\beta(\epsilon)^2}{2}  - c_4 C_P^2  &  - \frac{c_4 C_P (L_\rho + c_5 \gamma) }{2} \\
- \frac{c_4 C_P (L_\rho + c_5 \gamma) }{2} & \frac{c_3}{2} 
\end{bmatrix} \\
&  d(\sigma, \epsilon)  =  \sigma d_\eta(\epsilon)^2 + \left \lbrack \frac{2 C_P ^2 }{\sigma} + \frac{1}{c_4}  + \frac{(L_\rho + c_5 \gamma)^2}{2 c_3} \right \rbrack c_4^2 d_P(\epsilon)^2 . \nonumber
\end{align}
By choosing a large enough $\sigma$, it can be verified that $\Lambda_{\HybridSystem}$ is positive definite. We complete the proof by picking an even larger $\epsilon > \bar \epsilon$ such that $d(\sigma, \epsilon)$ is sufficiently small.% (to allow contraction of the level sets of $V$).
\end{pf}
\gap

\section{Results}
\label{sec:results}

In this section we will discuss simulation results on a $2$-link walker platform shown in \figref{fig:amber1}. %
%\subsection{AMBER1}
The biped is planar with $2$ links; this model was primarily used to establish robust walking behaviors in \cite{Russ_IJRR11}. The robot has point feet, and is thus underactuated at the ankle. Table \ref{tab:cg} provides the list of physical parameters of the $2$-link walker. %Masses of the links are $0.103$ kg and the mass of the hip is $0.068$ kg. 
More details about this model are found in \cite{Russ_IJRR11}.

\begin{figure}[!ht]\centering
\includegraphics[height=0.18\textwidth,align=t]{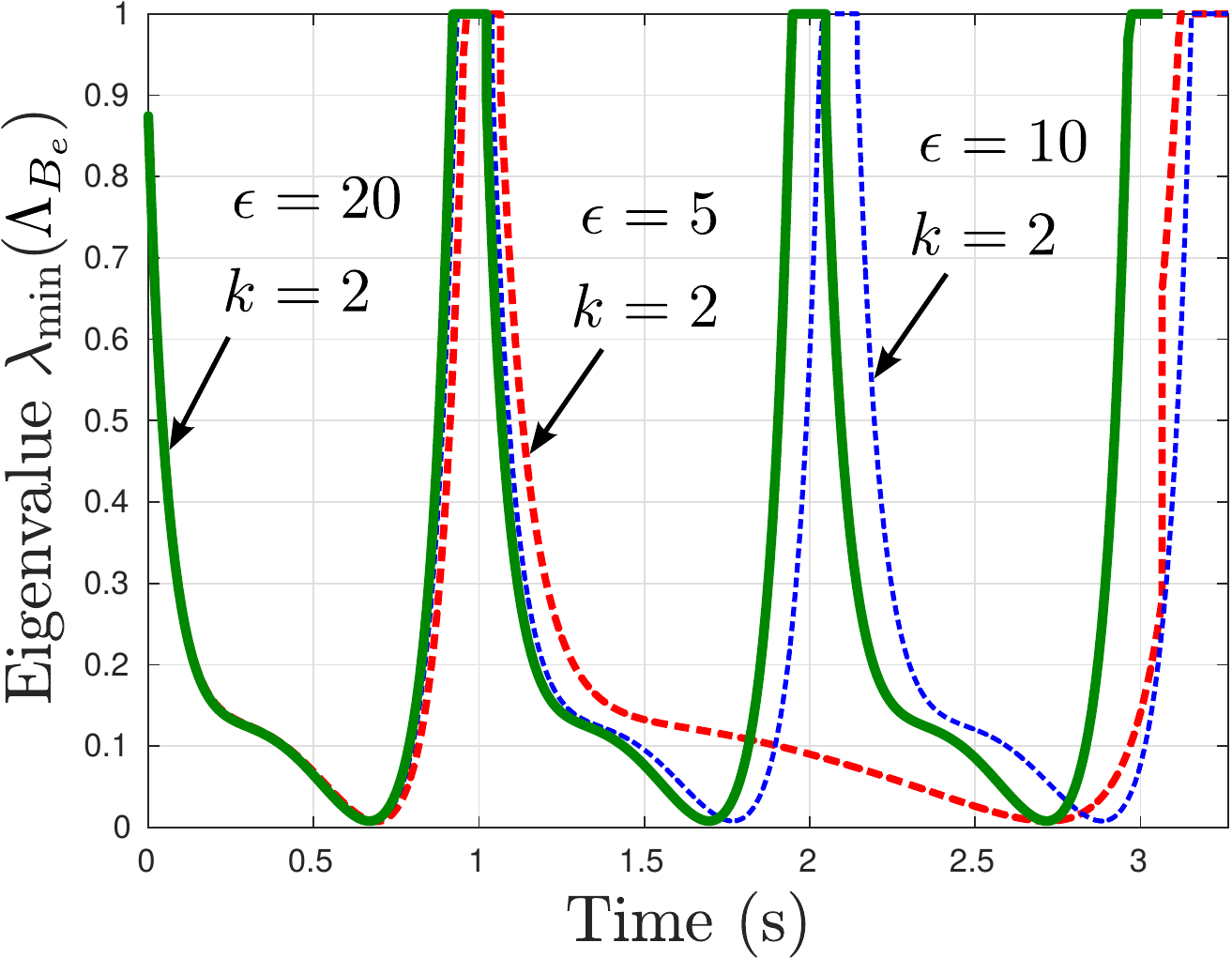} 
\includegraphics[height=0.18\textwidth,align=t]{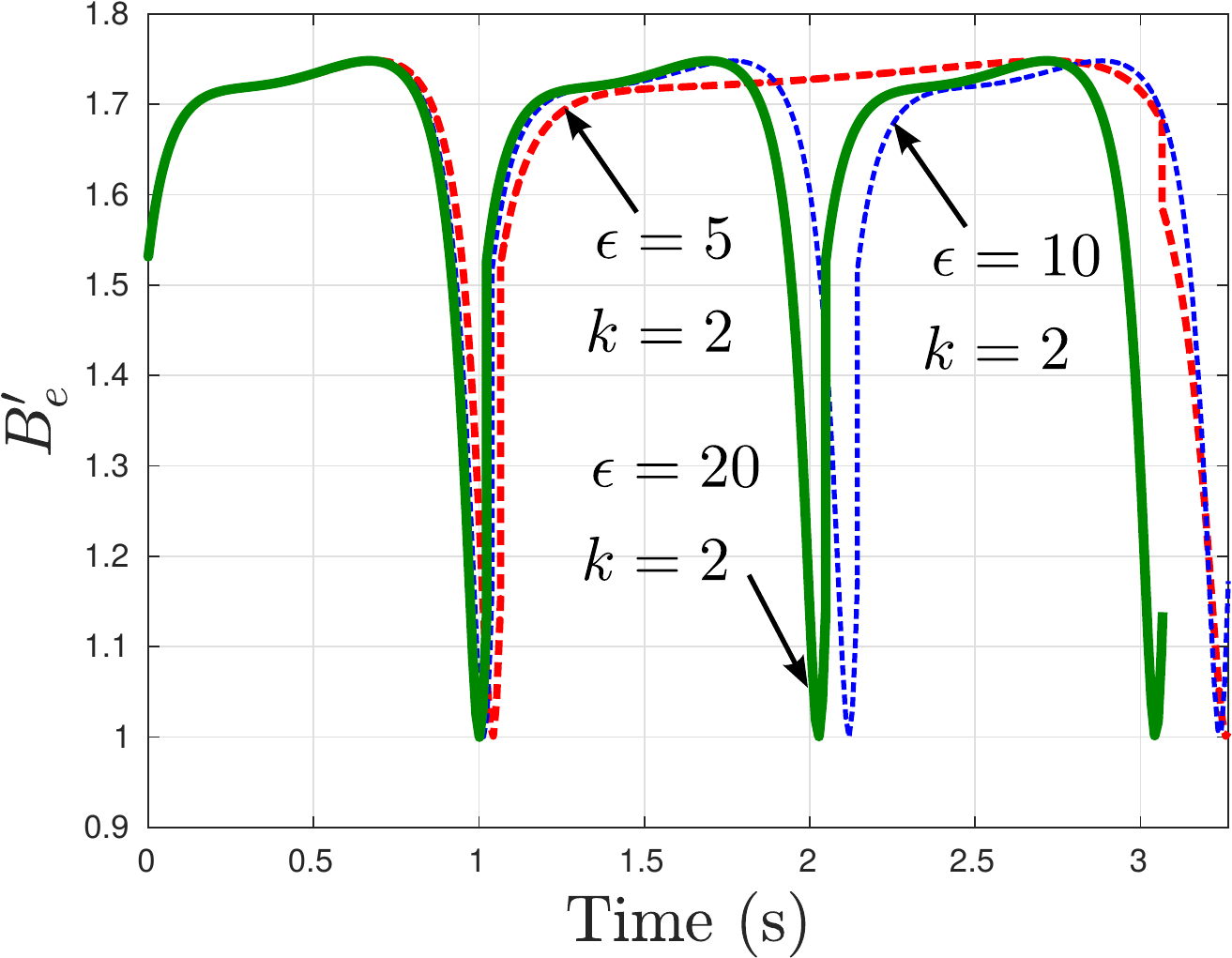} 
	\caption{Left figure is showing the minimum eigenvalue of $\Lambda_{B_e}$ (see Assumption \ref{ass:by}), and the right figure is showing   $B_e'$ for three steps. This shows that Assumption \ref{ass:by} is valid for the gait obtained. 
	}
	\label{fig:bebeprime}
\end{figure}

\begin{figure*}[!ht]
\centering \hspace{-2mm}
\includegraphics[height=0.215\textwidth,align=t]{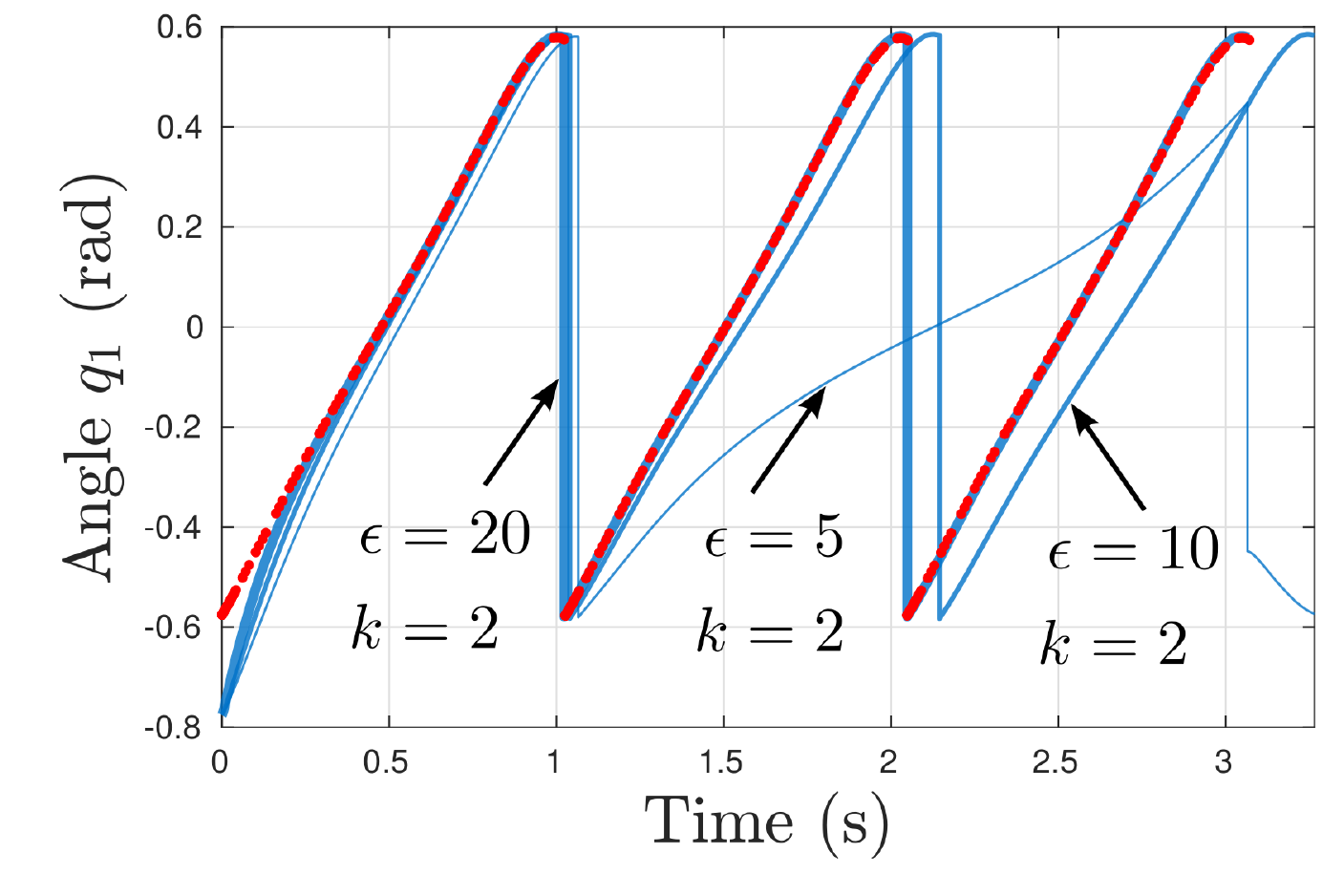} 
\includegraphics[height=0.21\textwidth,align=t]{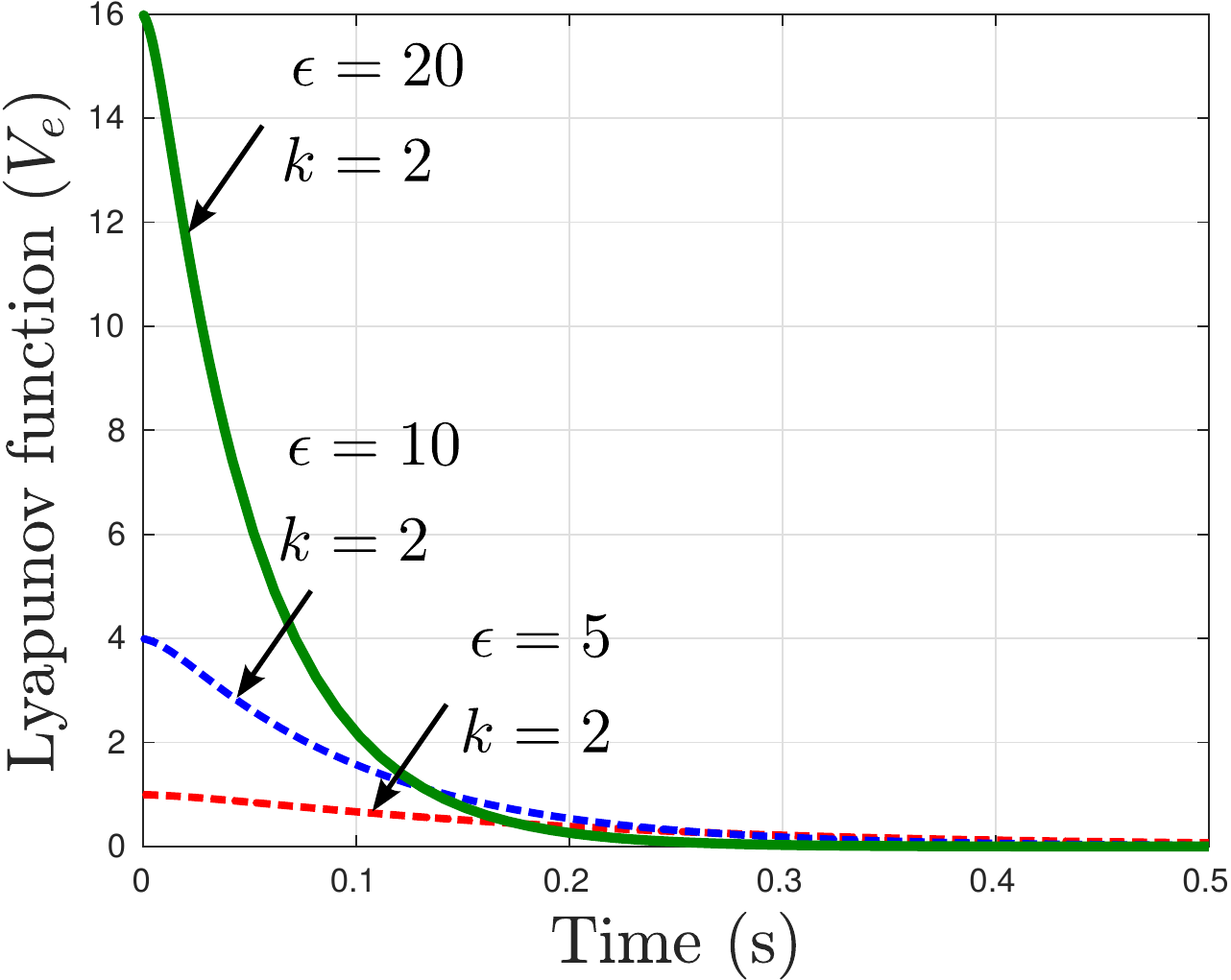} 
\includegraphics[height=0.21\textwidth,align=t]{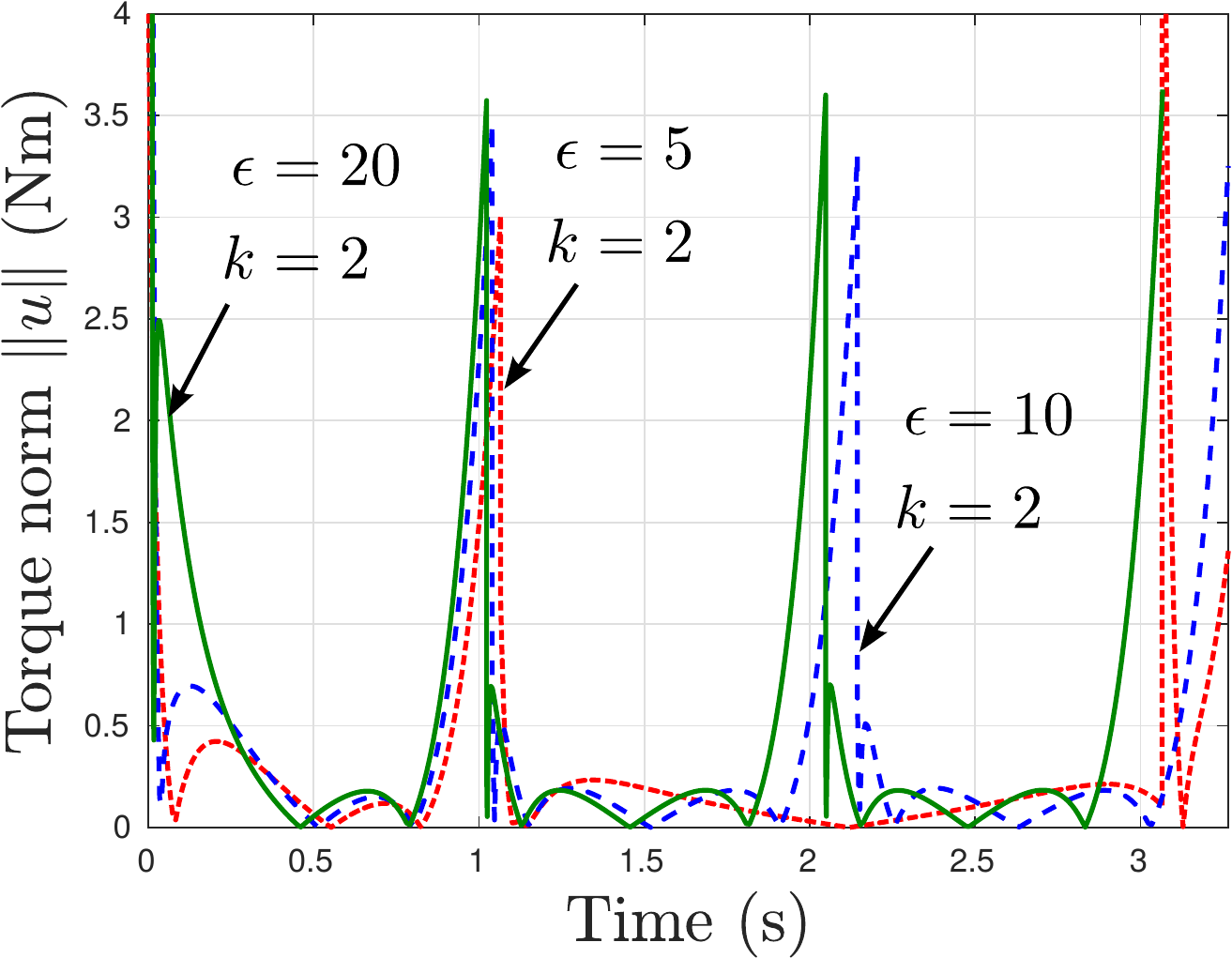} 
	\caption{Left figure is showing the comparison between the actual and nominal (dashed) trajectories for different gains of the control law \eqref{eq:pdtrackingunderactuated}.
	The center figure is showing the Lyapunov functions varying as a function of time. The right figure is showing the corresponding torque values varying as a function of time. Since the Lyapunov functions are decaying fast, their plots are shown for a shorter duration. Note that the nominal trajectories are not the same as the desired trajectories.
	}
	\label{fig:amber1trackingerror}
\end{figure*}

\begin{figure}[!ht]\centering 
\includegraphics[height=0.17\textwidth,align=t]{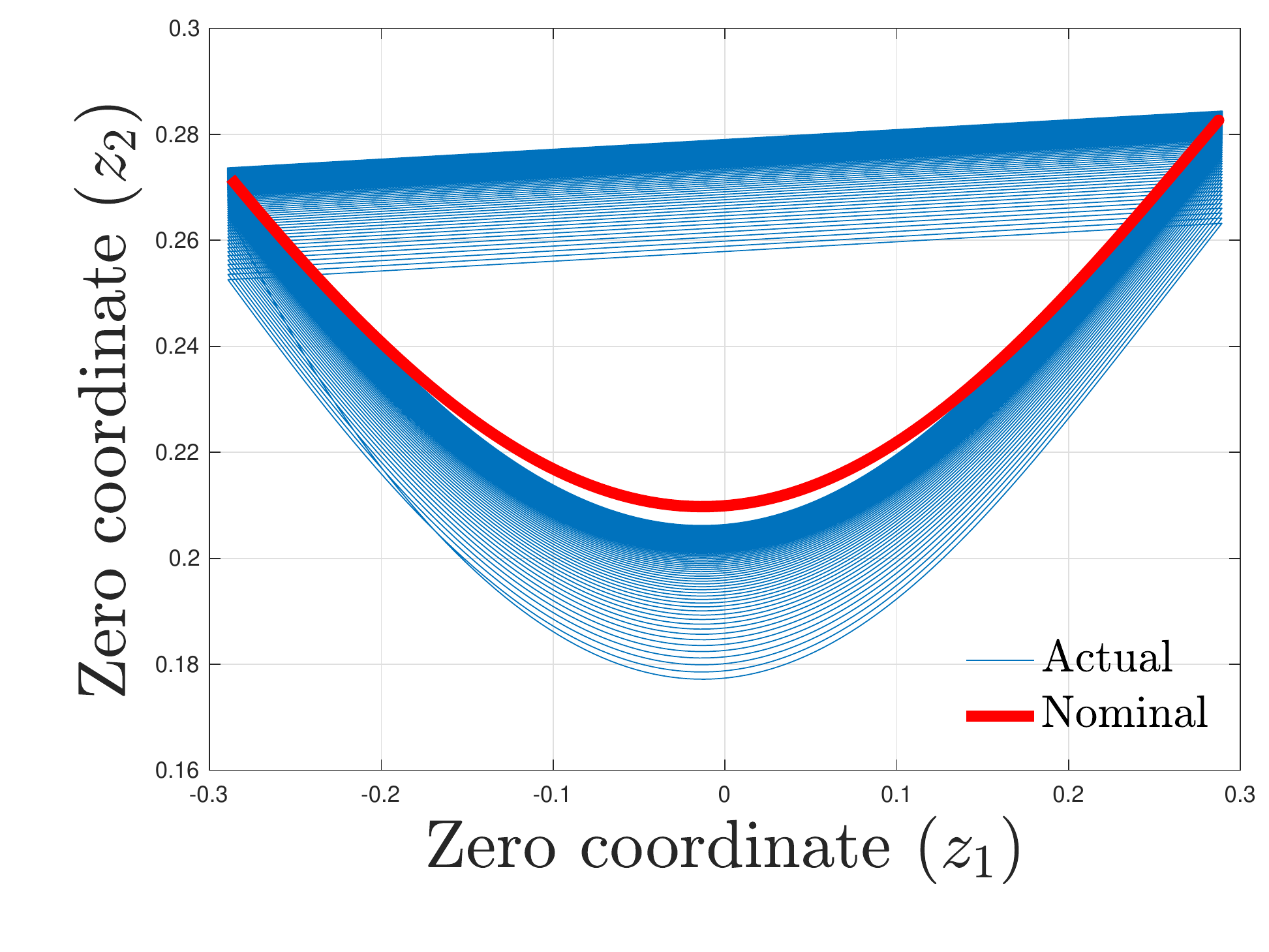} 
\includegraphics[height=0.17\textwidth,align=t]{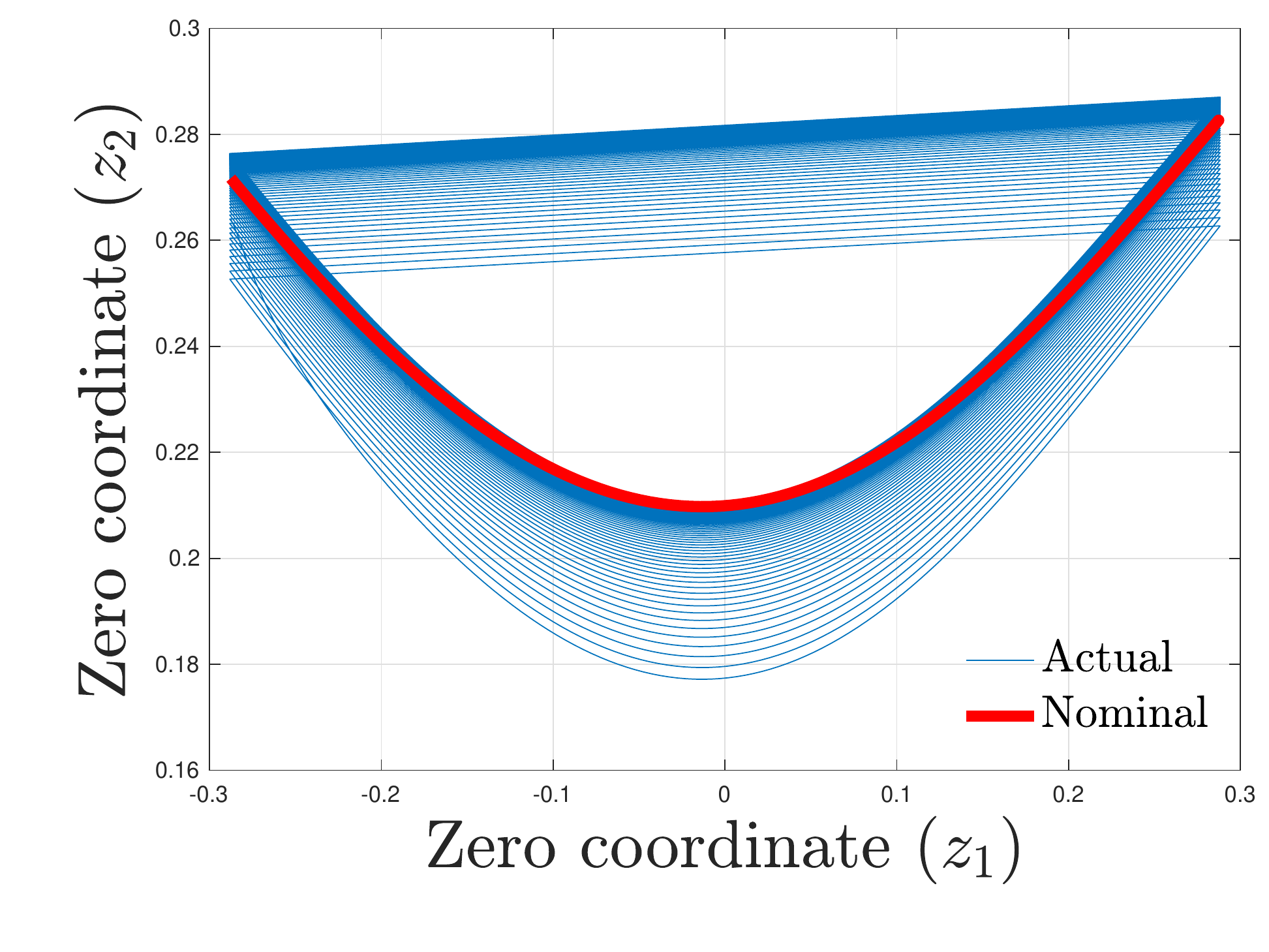} 
	\caption{Figure showing the phase portrait of the zero coordinates for $\epsilon = 20$ (left) and for $\epsilon = 40$ (right). The phase portrait for the nominal gait is shown for reference. The thick band indicates the region where the walking trajectories are mostly lying. The band is thinner and closer to the nominal curve for $\epsilon = 40$.}
	\label{fig:pp}
\end{figure}

\newsec{Outputs and control.} We have the configuration as
$	q = (q_{1}, q_{2}),$ where $q_{1}$ is the angle between the legs i.e., angle between the stance and nonstance legs, $q_{2}$ is the stance leg angle w.r.t. the vertical axis. The output is defined as follows:
\begin{align}
	e(q) = q_1  - q^a_d(\tau(q_2)),
\end{align}
where $q^a_d$ is a $5^{th}$ order polynomial:
\begin{align}
\label{eqn:smd}
q^a_d(\tau) = \sum\limits_{j=0}^{5} \zeta_{j} \tau^{5-j} (1 - \tau)^{j}, \quad \tau(q_2) = \frac{q_2 - q_0}{q_f - q_0},%+ \zeta_{1} q_2 +  \zeta_{2} q^2_2 + \zeta_{3} q^3_2 + \zeta_{4} q^4_2 + \zeta_{5} q^5_2,
\end{align}
with $\zeta_{j}$'s being the parameters $0.5753$, $3.1632$, $0.3115$, $-0.0570$, $-1.9988$, $-0.5753$ in increasing order, and $q_0, q_f$ being the initial and final values of $q_2$ on the orbit. These parameters are obtained via an offline optimization problem \cite{Hereid_etal_2016,Yadukumar2012a,Yadukumar2013}. The control law is given by \eqref{eq:pdtrackingunderactuated}. We chose the following gain values, and then studied the resulting walking:
\begin{align}
k_p = \epsilon^2, \quad k_d =  \epsilon k, \quad {\rm{with}} \quad \epsilon = 5,10,20, \quad k = 2.
\end{align}
$\epsilon = 5$ resulted in eventual falling, while $\epsilon = 10, 20$ resulted in stable walking. We verified that Assumption \ref{ass:by} was satisfied throughout the step, which is shown in Fig. \ref{fig:bebeprime}. Fig. \ref{fig:amber1trackingerror} shows the responses and torque profiles for these different gains. Fig. \ref{fig:pp} shows the phase portrait of the zero coordinates for $100$ steps for $\epsilon = 20, 40$.  It can be verified from Figs. \ref{fig:amber1trackingerror} and \ref{fig:pp} that increasing $\epsilon$ not only increases the convergence rate, but also reduces the ultimate bound. 

	\begin{table}[!ht]
\begin{center}
	\begin{tabular}{| c | c | c | c |}
		\hline
		\multicolumn{4}{| c |}{   {\small \textbf{Model Parameters}}    }\\ \hline
		{\it Link}         & {\it Mass}	 	& {\it Length}	 & {\it Center} 	 \\
		&	($kg$) &	$l$($m$)	& $l_c$ ($m$)	\\ \hline
		{\small Leg}  	& {\small $0.103$}	& {\small $0.5$}	 & {\small $0.33$}              \\ \hline
		{\small Hip}  	& {\small $0.068$}	& {\small N/A}	       & {\small $0$} \\ \hline
		\end{tabular}
	      \caption{The table of parameters for the $2$-link walker is shown here.}
		\label{tab:cg}
\end{center}
	\end{table}

\section{Conclusions}
We established that PD control laws are sufficient to realize locally stable periodic orbits in underactuated hybrid robotic systems. As an example, we have realized stable walking in a $2$-link bipedal walker. The key methodology is to use derivative gains that are at least as high as the square root of the proportional gains, and then increase the proportional gains as high as possible. It is important to note that properties like existence of stable periodic orbits in HZD, and low variations of the desired trajectories w.r.t. the unactuated coordinates have been used to achieve stable walking. We are also assuming that the joint actuators have high torque limits. These are not restrictive, and can be used as guidelines for the biped design process.

\begin{ack}
The author would like to acknowledge Sushant Veer and Wen-Loong Ma for providing useful insights and suggestions in improving the proofs in the paper.
\end{ack}

\appendix
 
\appendix

\section{Proofs of Propositions \ref{prop:schurpd} and \ref{prop:schurother}}
\label{FirstAppendix}
\begin{pf}\addblue{
	For convenience let $D_s (\q) := \Ba^T A(\q)D(\q) \Ba$. It is easy to see why $D_s$ is symmetric. By \cite[A.5.5]{boyd2004convex}, $D_s$ is positive definite.
	 We consider the inverse of $D$, which is obtained as
	\begin{align}\label{eq:Dinverseinschurform}
	D^{-1} = \begin{bmatrix} 
	D_{11}^{-1} + D_{11}^{-1} D_{12} D_s^{-1} D_{21} D_{11}^{-1} &  - D_{11}^{-1} D_{12} D_s^{-1} \\
	-D_s^{-1} D_{21} D_{11}^{-1}  & D_s^{-1}
	\end{bmatrix}.
	\end{align}
	%and since $D$ is positive definite, we know that its inverse $D^{-1}$ is also positive definite, and every block diagonal sub-matrix of $D^{-1}$ must be positive definite. Therefore $D_s$  must be positive definite. In addition, 
	Since $\|D^{-1}\|$ has upper and lower bounds (eigenvalues of $D^{-1}$ are the inverse eigenvalues of $D$), $D_s$ must have upper and lower bounds. We will choose these bounds to be $c_l, c_u$ respectively. The rest of the properties can be obtained directly from Property \ref{prop:1} (since $D$ contains only sine and cosine functions of $\q$, their derivatives are bounded w.r.t. $\q$).}
\end{pf}

\section{Proof of Proposition \ref{proposition:dece}}
\label{SecondAppendix}
\begin{pf}\addblue{
Positive definiteness of $D_e$, and skew symmetry of $\dot D_e  - 2 C_e$ are shown in \cite[Lemma 4.11]{MLS94}.
% 	It is easy to see why $D_e$ is symmetric positive definite. Further, we take the derivative of $D_e$ to yield
% 	\begin{align}
% 	\dot D_e = J_e^{-T} \dot D J_e^{-1}  + 2 J_e^{-T} D \frac{ d (J_e^{-1})}{dt},
% 	\end{align}
% 	and $\dot D_e - 2 C_e$ yields 
% 	\begin{align}
% 	\dot D_e - 2 C_e = J_e^{-T} (\dot D - 2 C) J_e^{-1},
% 	\end{align}
% 	which is skew-symmetric. 
	To prove the second part of the proposition, we have that
	\begin{align}\label{eq:desplit1}
	\| D_e \| \leq \| J_e^{-T} \| \| D \| \| J_e^{-1}\|, 
	\end{align}
	and based on boundedness and invertibility of $J_e$, we choose a $c_u$ (possibly larger) that bounds $D_e$. Similarly
	\begin{align}\label{eq:desplit2}
	\| D \| \leq \| J_e^{T} \| \| D_e \| \| J_e\|,
	\end{align}
	and the lower bound can be obtained accordingly. Similar procedure follows for $D_e^{-1}$,$G_e$. For $\dot D_e$ we have
	\begin{align}\label{eq:desplit3}
	 \| \dot D_e \| = \left \| \frac{\partial D_e}{\partial \q} J_e^{-1} \begin{bmatrix} \dq^u \\ \dot e \end{bmatrix} \right \| \leq c_u (|\dq^u| + |\dot e|),
	\end{align}
	for some $c_u >0$. Similar procedure follows for $C_e$.}
\end{pf}

\section{Proof of Proposition \ref{prop:schurpde}}
\label{ThirdAppendix}

\begin{pf}\addblue{
Proof is very similar to Proof of Propositions \ref{prop:schurpd}, \ref{prop:schurother}, and \ref{proposition:dece}. We note that $A_e, D_e$ are purely functions of the configuration $\q$ from a compact space $\mathbb{N}$. Therefore, the arguments follow directly from \eqref{eq:Dinverseinschurform}, \eqref{eq:desplit1}, \eqref{eq:desplit2}, \eqref{eq:desplit3}.}
\end{pf}

%%%%%%%%%%%%%%%%%%%%%%%%%%%%%%%%%%%%%%%%%%%%%%%%%%%%%%%%%%%%%%%%%%%%%%%%%%%%%%%%
\bibliographystyle{plain}
\bibliography{bibdata}

\begin{thebibliography}{10}

\bibitem{TAC:amesCLF}
A.~D. Ames, K.~Galloway, K.~Sreenath, and J.~W. Grizzle.
\newblock Rapidly exponentially stabilizing control {L}yapunov functions and
  hybrid zero dynamics.
\newblock {\em IEEE Transactions on Automatic Control}, 59(4):876--891, 2014.

\bibitem{ARIMOTO1985221}
S.~Arimoto and F.~Miyazaki.
\newblock Asymptotic stability of feedback control laws for robot manipulator.
\newblock {\em IFAC Proceedings Volumes}, 18(16):221 -- 226, 1985.
\newblock 1st IFAC Symposium on Robot Control.

\bibitem{boyd2004convex}
S.~Boyd and L.~Vandenberghe.
\newblock {\em Convex optimization}.
\newblock Cambridge university press, 2004.

\bibitem{brakken2007gershgorin}
S.~Brakken-Thal.
\newblock Gershgorin’s theorem for estimating eigenvalues.
\newblock 2007.

\bibitem{choi2004pid}
Y.~Choi and W.~K. Chung.
\newblock {\em PID trajectory tracking control for mechanical systems}, volume
  298.
\newblock Springer, 2004.

\bibitem{ROB:ROB2}
F.~Ghorbel, B.~Srinivasan, and M.~W. Spong.
\newblock On the uniform boundedness of the inertia matrix of serial robot
  manipulators.
\newblock {\em Journal of Robotic Systems}, 15(1):17--28, 1998.

\bibitem{grizzle20103d}
J.~W. Grizzle, Christine Chevallereau, Aaron~D. Ames, and Ryan~W. Sinnet.
\newblock 3d bipedal robotic walking: Models, feedback control, and open
  problems.
\newblock {\em IFAC Proceedings Volumes}, 43(14):505 -- 532, 2010.

\bibitem{umich_mabel}
J.~W. Grizzle, J.~Hurst, B.~Morris, H.~Park, and K.~Sreenath.
\newblock {MABEL}, a new robotic bipedal walker and runner.
\newblock In {\em American Control Conference}, pages 2030--2036, St. Louis,
  MO, USA, 2009.

\bibitem{gravityboundedness}
R.~{Gunawardana} and F.~{Ghorbel}.
\newblock The class of robot manipulators with bounded jacobian of the gravity
  vector.
\newblock In {\em Proceedings of IEEE International Conference on Robotics and
  Automation}, volume~4, pages 3677--3682 vol.4, April 1996.

\bibitem{kaveh_expo_ijrr}
K.~A. Hamed, B.~G. Buss, and J.~W. Grizzle.
\newblock Exponentially stabilizing continuous-time controllers for periodic
  orbits of hybrid systems: Application to bipedal locomotion with ground
  height variations.
\newblock {\em The International Journal of Robotics Research}, 35(8):977--999,
  2016.

\bibitem{Hereid_etal_2016}
A.~Hereid, E.~A. Cousineau, C.~M. Hubicki, and A.~D. Ames.
\newblock 3d dynamic walking with underactuated humanoid robots: A direct
  collocation framework for optimizing hybrid zero dynamics.
\newblock In {\em 2016 IEEE International Conference on Robotics and Automation
  (ICRA)}, pages 1447--1454, 5 2016.

\bibitem{Hubicki2016}
C.~Hubicki, J.~Grimes, M.~Jones, D.~Renjewski, A.~Spr{\"o}witz, A.~Abate, and
  J.~Hurst.
\newblock Atrias: Design and validation of a tether-free {3D}-capable
  spring-mass bipedal robot.
\newblock {\em The International Journal of Robotics Research}, 2016.

\bibitem{hurmuzlu:biped}
Y.~Hurmuzlu.
\newblock Dynamics of bipedal gait; part i: Objective functions and the contact
  event of a planar five-link biped.
\newblock {\em ASME Journal of Applied Mechanics}, 60(2):331--336, 1993.

\bibitem{hurmuzlu_marghitu}
Y.~Hurmuzlu and D.~B. Marghitu.
\newblock Rigid body collisions of planar kinematic chains with multiple
  contact points.
\newblock {\em The International Journal of Robotics Research}, 13(1):82--92,
  1994.

\bibitem{12253}
S.~Kawamura, F.~Miyazaki, and S.~Arimoto.
\newblock Is a local linear pd feedback control law effective for trajectory
  tracking of robot motion?
\newblock In {\em Proceedings of IEEE International Conference on Robotics and
  Automation}, pages 1335--1340 vol.3, April 1988.

\bibitem{khalil2002nonlinear}
H.~K. Khalil.
\newblock {\em Nonlinear systems}, volume~3.
\newblock Prentice hall Upper Saddle River, NJ, 2002.

\bibitem{koditschek1984natural}
D.~E. Koditschek.
\newblock Natural motion for robot arms.
\newblock In {\em Proceedings of IEEE Conference on Decision and Control},
  volume~23, pages 733--735. IEEE, 1984.

\bibitem{koditschek1987adaptive}
D.~E. Koditschek.
\newblock Adaptive techniques for mechanical systems.
\newblock In {\em Fifth Yale Workshop on Applications of Adaptive Systems
  Theory, May 1987}, pages 259--265, 1987.

\bibitem{koditschek1987highgain}
D.~E. Koditschek.
\newblock High gain feedback and telerobotic tracking.
\newblock In {\em Proceedings of the Workshop on Space Telerobotics}, pages
  355--363, 1987.

\bibitem{koditschek1988global}
D.~E. Koditschek.
\newblock Strict global lyapunov functions for mechanical systems.
\newblock In {\em Proceedings of American Control Conference}, pages
  1770--1775, June 1988.

\bibitem{koditschek-the_robotics_review-1989}
D.~E. Koditschek.
\newblock Robot planning and control via potential functions.
\newblock pages 349--367, 1989.

\bibitem{kolathaya2015parameter}
S.~Kolathaya and A.~D. Ames.
\newblock Parameter sensitivity and boundedness of robotic hybrid periodic
  orbits.
\newblock {\em IFAC-PapersOnLine}, 48(27):377 -- 382, 2015.

\bibitem{kolathaya2016parameter}
S.~Kolathaya and A.~D. Ames.
\newblock Parameter to state stability of control {L}yapunov functions for
  hybrid system models of robots.
\newblock {\em Nonlinear Analysis: Hybrid Systems}, 25:174 -- 191, 2017.

\bibitem{kolathaya2016time}
S.~Kolathaya, A.~Hereid, and A.~D. Ames.
\newblock Time dependent control {L}yapunov functions and hybrid zero dynamics
  for stable robotic locomotion.
\newblock In {\em 2016 American Control Conference (ACC)}, pages 3916--3921, 7
  2016.

\bibitem{LUinverseblockmatrix}
T-T. Lu and S-H. Shiou.
\newblock Inverses of 2 x 2 block matrices.
\newblock {\em Computers \& Mathematics with Applications}, 43(1):119 -- 129,
  2002.

\bibitem{hscc17running}
W-L. Ma, S.~Kolathaya, E.~R. Ambrose, C.~M. Hubicki, and A~D. Ames.
\newblock Bipedal robotic running with {DURUS-2D}: Bridging the gap between
  theory and experiment.
\newblock In {\em Proceedings of the 20th International Conference on Hybrid
  Systems: Computation and Control}, HSCC '17, pages 265--274, NY, USA, 2017.
  ACM.

\bibitem{Russ_IJRR11}
I.~R. Manchester, U.~Mettin, F.~Iida, and R.~Tedrake.
\newblock Stable dynamic walking over uneven terrain.
\newblock {\em The International Journal of Robotics Research}, 30(3):265--279,
  2011.

\bibitem{MOGR05}
{B.} Morris and {J. W.} Grizzle.
\newblock A restricted {P}oincar\'e map for determining exponentially stable
  periodic orbits in systems with impulse effects: Application to bipedal
  robots.
\newblock In {\em IEEE Conf. on Decision and Control}, Seville, Spain, 2005.

\bibitem{4339540}
J.~I. Mulero-Martinez.
\newblock Uniform bounds of the {C}oriolis/centripetal matrix of serial robot
  manipulators.
\newblock {\em Robotics, IEEE Transactions on}, 23(5):1083--1089, 10 2007.

\bibitem{MLS94}
R.~M. Murray, {Zexiang} Li, and S.~S. Sastry.
\newblock {\em A Mathematical Introduction to Robotic Manipulation}.
\newblock CRC Press, Boca Raton, 1994.

\bibitem{posa2014direct}
M.~Posa, C.~Cantu, and R.~Tedrake.
\newblock A direct method for trajectory optimization of rigid bodies through
  contact.
\newblock {\em The International Journal of Robotics Research}, 33(1):69--81,
  2014.

\bibitem{reheralgorithmic}
J.~P. Reher, A.~Hereid, S.~Kolathaya, C.~M. Hubicki, and A.~D. Ames.
\newblock {\em Algorithmic Foundations of Realizing Multi-Contact Locomotion on
  the Humanoid Robot {DURUS}}.
\newblock Springer Berlin Heidelberg, Berlin, Heidelberg, 2017.

\bibitem{7823045}
T.~Samad.
\newblock A survey on industry impact and challenges thereof [technical
  activities].
\newblock {\em IEEE Control Systems}, 37(1):17--18, Feb 2017.

\bibitem{takegaki1981new}
M.~Takegaki and S.~Arimoto.
\newblock {A New Feedback Method for Dynamic Control of Manipulators}.
\newblock {\em Journal of Dynamic Systems Measurement and Control-transactions
  of The Asme}, 103, 1981.

\bibitem{veer2017poincare}
S.~{Veer}, {Rakesh}, and I.~{Poulakakis}.
\newblock Input-to-state stability of periodic orbits of systems with impulse
  effects via poincaré analysis.
\newblock {\em IEEE Transactions on Automatic Control}, 64(11):4583--4598, Nov
  2019.

\bibitem{wen1988new}
J.~T. Wen and D.~S. Bayard.
\newblock New class of control laws for robotic manipulators part 1.
  non--adaptive case.
\newblock {\em International Journal of Control}, 47(5):1361--1385, 1988.

\bibitem{WGK03}
E.~R. Westervelt, J.~W. Grizzle, and D.~E. Koditschek.
\newblock Hybrid zero dynamics of planar biped walkers.
\newblock {\em IEEE Transactions on Automatic Control}, 48(1):42--56, 2003.

\bibitem{whitcombpd1993}
L.~L. Whitcomb, A.~A. Rizzi, and D.~E. Koditschek.
\newblock Comparative experiments with a new adaptive controller for robot
  arms.
\newblock {\em IEEE Transactions on Robotics and Automation}, 9(1):59--70, Feb
  1993.

\bibitem{Yadukumar2012a}
S.~N. Yadukumar, M.~Pasupuleti, and A.~D. Ames.
\newblock Human-inspired underactuated bipedal robotic walking with {AMBER} on
  flat-ground, up-slope and uneven terrain.
\newblock In {\em 2012 IEEE/RSJ International Conference on Intelligent Robots
  and Systems}, pages 2478--2483, October 2012.

\bibitem{Yadukumar2013}
S.~N. Yadukumar, M.~Pasupuleti, and A.~D. Ames.
\newblock {\em From Formal Methods to Algorithmic Implementation of Human
  Inspired Control on Bipedal Robots}, pages 511--526.
\newblock Springer Berlin Heidelberg, Berlin, Heidelberg, 2013.

\end{thebibliography}

\begin{wrapfigure}{l}{2.6cm}
\vspace{-3mm}
\includegraphics[width=0.175\textwidth]{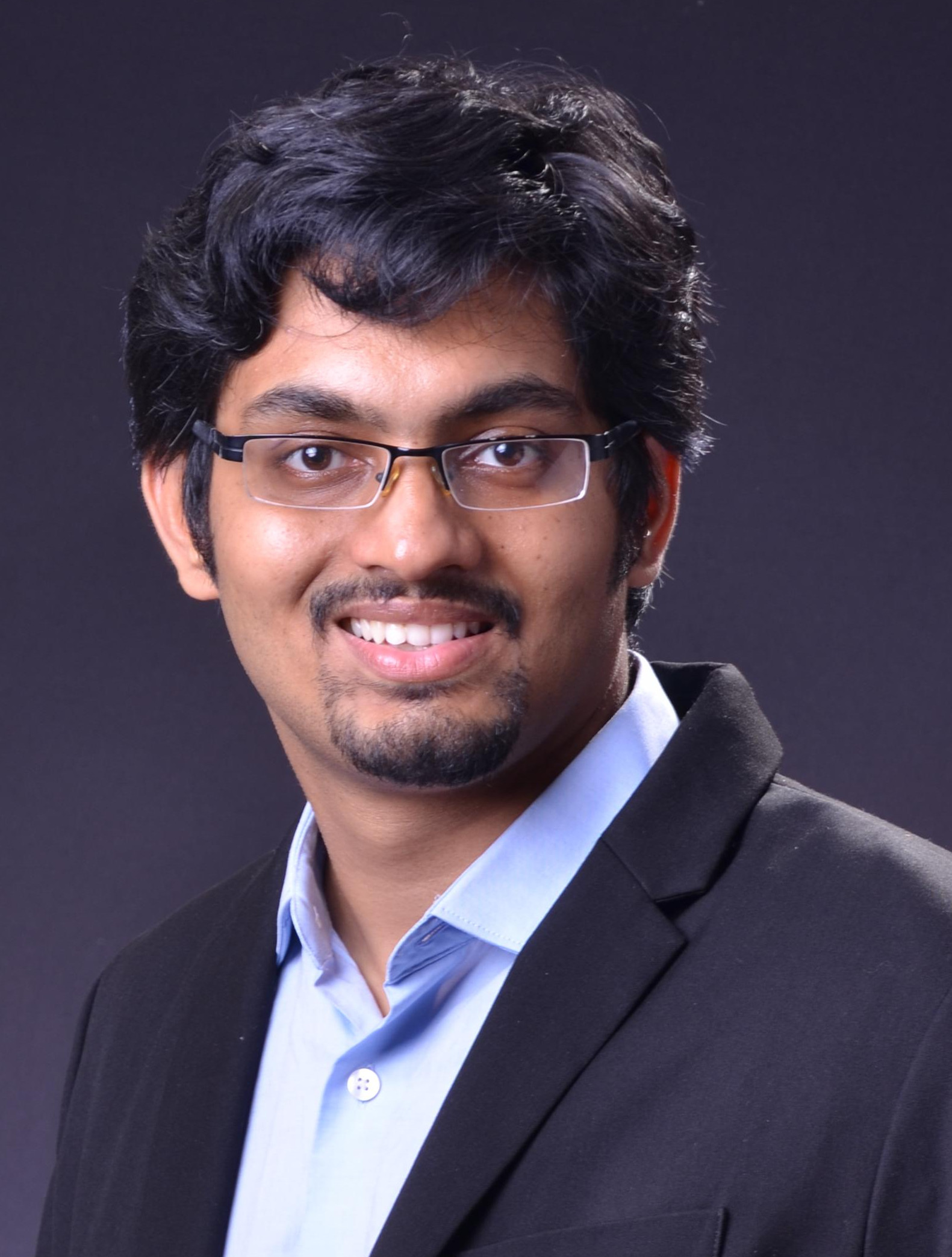}
\end{wrapfigure}
\vspace{1mm}
\noindent \textbf{Shishir Kolathaya} is an INSPIRE Faculty Fellow in the Robert Bosch Center for Cyber Physical Systems at the Indian Institute of Science, Bengaluru, India. Previously, he was a Postdoctoral Scholar in the department of Mechanical and Civil Engineering at the California Institute of Technology. He received his Ph.D. degree in Mechanical Engineering (2016) from the Georgia Institute of Technology, his M.S. degree in Electrical Engineering (2012) from Texas A\&M University, and his B. Tech. degree in Electrical \& Electronics Engineering (2008) from the National Institute of Technology Karnataka, Surathkal. Shishir is interested in stability and control of nonlinear hybrid systems, especially in the domain of legged robots. His more recent work is also focused on real-time safety-critical control for robotic systems.

\end{document}